\documentclass{article}

\PassOptionsToPackage{numbers, compress}{natbib}

 \usepackage[main, final]{neurips_2026}

\usepackage[utf8]{inputenc} 
\usepackage[T1]{fontenc}    
\usepackage{hyperref}       
\usepackage{url}            
\usepackage{booktabs}       
\usepackage{amsfonts}       
\usepackage{nicefrac}       
\usepackage{microtype}      
\usepackage{xcolor}         
\usepackage{amsmath}
\usepackage{amsthm}
\usepackage{graphicx}
\usepackage{xcolor}
\usepackage{soul}
\usepackage{multirow}
\usepackage{makecell}
\usepackage{subcaption}
\usepackage{wrapfig}

\newtheorem{proposition}{Proposition}

\definecolor{rjgreen}{RGB}{70,220,50}
\definecolor{rjmid}{RGB}{170,220,50}

\newcommand{\ulbad}[1]{%
  {\setulcolor{red}\setul{1.2pt}{1.5pt}\ul{#1}}%
}
\newcommand{\ulgood}[1]{%
  {\setulcolor{rjgreen}\setul{1.2pt}{1.5pt}\ul{#1}}%
}

\newcommand{\ulmid}[1]{%
  {\setulcolor{rjmid}\setul{1.2pt}{1.5pt}\ul{#1}}%
}

\title{Early Signals, Strong Decisions: Prefix-Guided Sampling for Parallel Test-Time Scaling}

\author{%
  Jie Ren \\
  MIT CSAIL, MIT FutureTech \\
  \texttt{rjie@mit.edu} \\
  \And
  Jonathan Rosenfeld \\
  MIT CSAIL, MIT FutureTech \\
  \texttt{jonsr@csail.mit.edu} \\
  \And
  Neil Thompson \\
  MIT CSAIL, MIT FutureTech \\
  \texttt{neil\_t@mit.edu}
}

\begin{document}

\maketitle

\begin{abstract}

Parallel test-time scaling (TTS) has been shown to enhance reasoning in large language models by generating and aggregating multiple independent reasoning paths. Internal confidence scores can further improve this process by identifying paths of varying quality. However, existing methods remain computationally inefficient, as low-quality paths are fully expanded before being evaluated. In this work, we aim to improve the use of internal confidence signals for better model performance and computational efficiency.
We begin by offering a new perspective on the role of confidence in aggregation, showing that it is most impactful for questions near the decision boundary. Notably, when amplified by multi-sample aggregation, even small confidence signals derived from short prefixes can meaningfully influence outcomes in these boundary cases.
Building on this insight, we introduce Prefix-Guided Sampling (PreG), a method that reallocates computation by first generating short prefixes, ranking them using internal confidence scores, and then completing only the most promising candidates. This strategy reduces token usage while maintaining gains in path quality. We provide theoretical analysis demonstrating that, under a fixed compute budget, PreG strengthens the effective signal by widening the decision boundary margin. Empirically, our method consistently outperforms standard parallel TTS approaches across benchmarks, while significantly improving token efficiency and reducing overall computation. Our code and dataset is avaiable at \href{https://github.com/renjie3/PreG}{github.com/renjie3/PreG}.

\end{abstract}

\section{Introduction}


Large language models (LLMs) have demonstrated remarkable reasoning capabilities, particularly when augmented with test-time scaling (TTS) techniques that allocate additional computation during inference~\cite{wei2022chain, wangself, jaech2024openai, achiam2023gpt, huang2023towards, muennighoff2025s1}. Among these, parallel scaling has emerged as a powerful paradigm~\cite{wangself, brown2024large, snell2024scaling, wu2025inference}: multiple independent reasoning paths are sampled for the same problem, each producing an answer, and the answers are aggregated via mechanisms such as majority voting~\cite{wangself}, leading to improved performance.

These methods can be further improved by estimating the confidence of each path and assigning greater weight to high-confidence paths. Recent work has explored various confidence signals, including external supervision such as process reward models (PRMs)~\cite{lightman2023let, wang2024math} and verifier models~\cite{cobbe2021training, kim2026parallel}, as well as internal confidence signals~\cite{fu2025deep, kangscalable, fu2024efficiently, korikov2025batched, leang2025picsar, chen2026think, zhoutheoretical, zhang2025tokur}. The latter is particularly appealing because it requires no additional computation and generalizes across models. 
For example, DeepConf (DC) uses the concentration of prediction distributions across tokens~\cite{fu2025deep}, while Self-Certainty (SCert) utilizes the divergence between vocabulary distributions and a uniform prior at each token position~\cite{kangscalable}.

Despite the effectiveness of internal signal methods, these approaches suffer from a critical inefficiency~\cite{chen2025matryoshkathinking, wang2025sampling}. Each reasoning path is typically generated to completion before its confidence is assessed, 
as suggested in prior studies that confidence signals become more reliable in the later tokens of reasoning, closer to the final answer~\cite{fu2025deep, jin2026corefine}. This inefficiency becomes increasingly pronounced 
\begin{wrapfigure}{r}{0.34\linewidth}
    \vspace{-3pt}
    \centering
    \includegraphics[width=\linewidth]{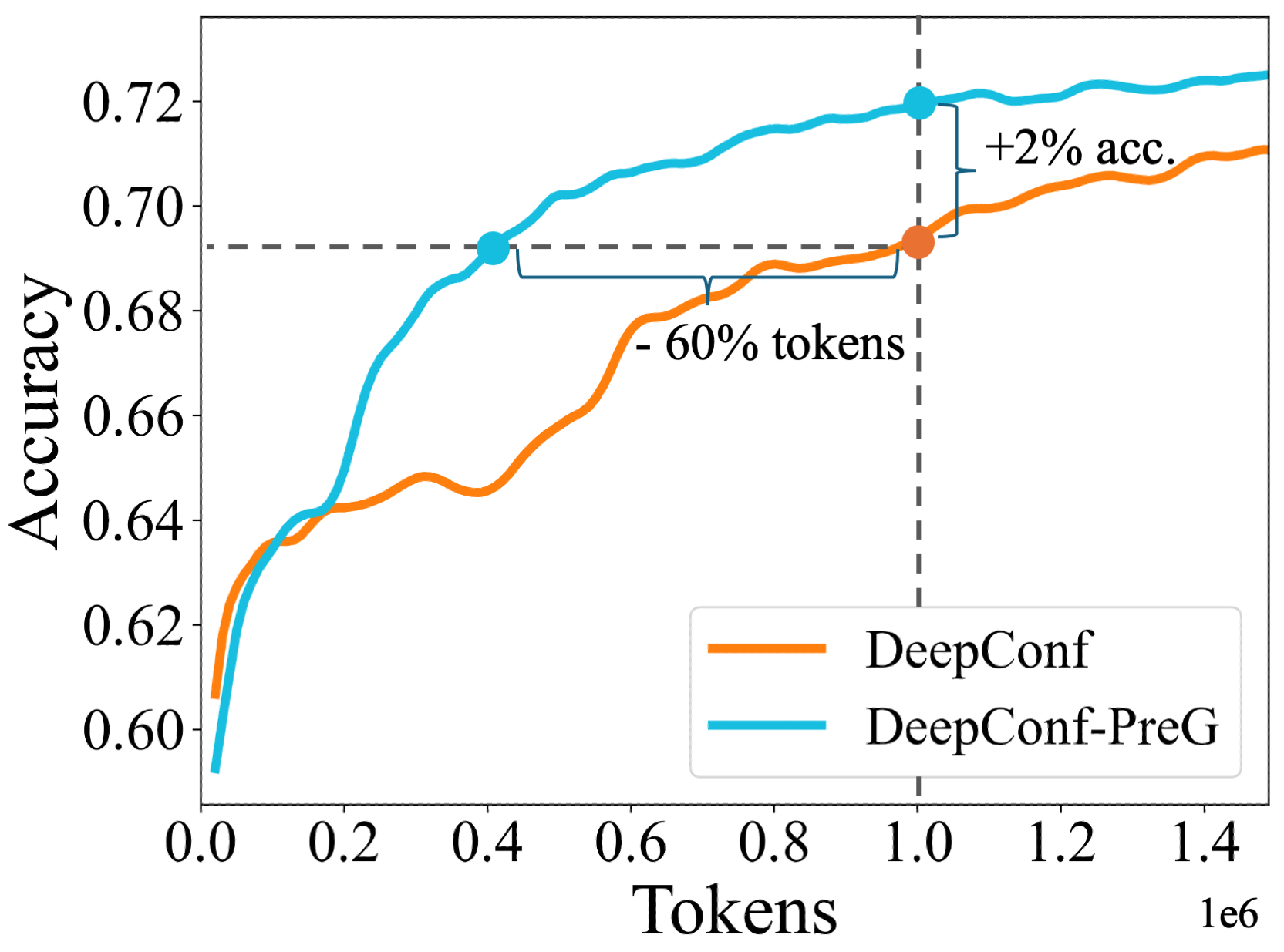}
    \vspace{-15pt}
    \caption{DC-PreG vs. vanilla DC}
    \label{fig:intro}
    \vspace{-20pt}
\end{wrapfigure}
as the number of paths grows, leading to substantial redundant token generation without proportional performance gains.

This raises a fundamental question:
\textit{Can we predict the confidence of a reasoning path from its prefix, before it is fully generated?}
If feasible, such an approach would substantially reduce the cost of identifying unpromising paths and improve the quality of the final path set. However, this problem is inherently challenging. Early-stage tokens are often incomplete and ambiguous, making prefix-based confidence scores unreliable. As a result, this direction has received limited attention~\cite{fu2025deep}.


In contrast to the common view that prefixes are uninformative, we show that even limited prefix information can be effectively leveraged when amplified through the aggregation process in parallel TTS. To verify this, we first provide a deeper understanding of the interaction between confidence scores and aggregation. Aggregation enhances the most probable answer: when correct paths dominate, it reinforces the right answer; when incorrect paths dominate, it amplifies the error. If there is a large gap between the numbers of correct and incorrect paths, confidence scores have little impact. But \underline{when counts are similar, confidence becomes critical}—even a slight boost to correct answers gets amplified through aggregation. This can accelerate convergence and even reverse the direction of amplification from incorrect to correct paths. Thus, even weak signals may suffice to influence the final outcome through aggregation.

This insight motivates us to ask whether prefix signals, despite their incompleteness, can serve as a useful signal after amplification. Surprisingly, we found that even very short prefixes, covering only around 20\% of the full reasoning path, already contain sufficient signal to influence uncertain samples (Section~\ref{sec:preliminary}), suggesting that early signals may be far more actionable than previously believed. Under a fixed token budget, this is even more efficient than full generation.
Building on this, we propose \textbf{Prefix-Guided Sampling} (PreG), a compute-efficient framework that reallocates computation during TTS. Instead of expanding all sampled paths to completion, PreG first allocates part of the budget to generate short prefixes, estimates their potential by internal confidence, and selectively continues generation only for a subset of candidates. This approach enables early identification of low-quality paths, reducing unnecessary computation while preserving strong candidates.

Through theoretical analysis, we show that allocating tokens to prefixes remains highly worthwhile, even with fewer full reasoning paths. This makes it an effective and principled mechanism for reallocating compute. Empirically, as shown in Figure~\ref{fig:intro}, our method consistently outperforms vanilla confidence-based parallel TTS approaches~\cite{fu2025deep, kangscalable} in reasoning tasks~\cite{dekoninck2026matharena} and achieves substantial gains in token efficiency, reducing token usage by a large margin. Overall, PreG demonstrates that leveraging early prefix signals is not only feasible but also a fundamentally more efficient paradigm for test-time scaling.

\section{Related Works}

\textbf{Test-time scaling.} Test-time scaling has emerged as a powerful paradigm for enhancing LLM reasoning by allocating additional computation at inference time, without modifying model parameters \cite{wei2022chain, wangself, jaech2024openai, achiam2023gpt, huang2023towards}. The methods broadly fall into two complementary categories: sequential scaling and parallel scaling. Sequential scaling extends computation along the depth dimension. by generating longer, iteratively refined reasoning chains. Chain-of-Thought prompting \cite{wei2022chain} generates longer by eliciting step-by-step reasoning, while other works like Self-Refine \cite{madaan2023self} encouraged models to revise their outputs through internal feedback. Parallel scaling improves performance along the width dimension by sampling multiple independent reasoning paths and aggregating their outputs~\cite{wangself, brown2024large, snell2024scaling, wu2025inference}.

\textbf{Internal confidence.} Internal confidence is increasingly used to improve reasoning. A simple source is P(True), obtained by directly querying the model~\cite{xie2024calibrating, yang2026believe}. Beyond this, more stable estimates aggregate over the full token sequence, using metrics such as likelihood~\cite{taubenfeld2025confidence}, Gini Impurity~\cite{loh2011classification, kangscalable}, entropy~\cite{sharma2025sequential}, DC~\cite{fu2025deep}, SCert~\cite{kangscalable}, Tokur~\cite{zhang2025tokur} and DTR~\cite{chen2026think}. In addition, CoRefine proposes treating the token confidence sequence as time-series data and training a classifier on top of it~\cite{jin2026corefine}.



\textbf{Efficient TTS. }
Prior work improves TTS efficiency via early stopping, either based on the generated paths themselves---such as ASC~\cite{aggarwal2023let}, ESC~\cite{liescape}, and CGES~\cite{aghazadeh2025cges}---or through inter-path comparison, where methods like Slim-SC~\cite{hong2025slim} and~\cite{wang2025sampling} exploit latent consistency among early paths to prune candidates. However, these approaches require sometimes partially sequential sampling or dynamic decision-making, limiting system-level scalability, and often sacrifice accuracy in the process.




\section{When Confidence Scores Matter: Aggregation Amplifies Small Gains}
\label{sec:preliminary}

In this section, we first formalize the problem, then discuss how the role of confidence scores and why prefix confidences serve as effective confidence estimates.

\subsection{Problem Statement and Definitions}

Given an input problem \(Q\) and an LLM, we consider solving the problem by parallel test-time scaling methods that independently generate a set of reasoning paths under a fixed token budget $B$. 

\textbf{Parallel testing-time scaling.} Given $Q$, we define a path generated by the model as 
$
S = (s_1, s_2, \dots, s_T),
$
where each token \(s_t \in \mathcal{V}\), and \(T\) denotes the sequence length.
Under a fixed token budget \(B\), a parallel TTS method produces a set of reasoning paths
$
\mathcal{S}_B = \{S_1, S_2, \dots, S_M\},
$
where \(M\) is the number of complete path and depends on how the budget is allocated. (If all the budget is used for full path generation, we denote the path number as $
M_{\mathrm{base}} = {B} / {T}
$.) Each \underline{complete} path yields an answer \(y_i\), and a final answer is obtained via an aggregation function:
$
\hat{y} = A(y_1, y_2, \dots, y_M).
$
Let \(y^*\) denote the ground-truth answer.
The objective is to maximize the probability of correct prediction (i.e., $\hat{y} = y^*$) under a fixed token budget \(B\).


\textbf{Confidence-based parallel TTS.} In this work, we focus on internal confidence scores without external models or verifiers, and aggregate answers via weighted voting:
\[
\hat{y} = \arg\max_y W(S_i) \sum_{i=1}^{M} \mathbf{1}[y_i = y],
\]
where $W(\cdot)$ is the confidence-based aggregation function. \textbf{Self-Consistency} (SC)~\cite{wangself} is a special case where $W(\cdot) = 1$ (i.e. unweighted majority voting), which is the most foundamental method.


\textbf{Margin.}
By denoting the aggregated score as $\mathrm{Conf}(y) = W(S_i)\sum_{i=1}^{M}\mathbf{1}[y_i=y]$ and the strongest incorrect answer as $y_w = \arg\max_{y\neq y^*}\mathrm{Conf}(y)$, we define the margin as
\begin{equation}
\Delta = \mathrm{Conf}(y^*) - \mathrm{Conf}(y_w) = \sum_{i=1}^M Z_i,
\end{equation}
where $Z_i = W(S_i)\bigl(\mathbf{1}[y_i=y^*]-\mathbf{1}[y_i=y_w]\bigr)$ is the per-path contribution. The prediction is correct iff $\Delta > 0$:
$
\Pr(\hat y = y^*) = \Pr(\Delta > 0).
$

\subsection{A Margin-Based Analysis of Aggregation in Parallel TTS}



In this subsection, we first investigate what factors influence the final aggregated accuracy.
The above margin decomposition $\Delta = \sum_{i=1}^M Z_i$ reveals that the aggregated outcome in parallel TTS depends on how per-path contributions accumulate. To characterize this process, we analyze how prediction accuracy relates to the statistical properties of these contributions, summarized in the following theorem which serves as the theoretical foundation for our subsequent analysis.

\begin{proposition}[Accuracy via margin]
\label{thm:margin-accuracy}
Assume that $\{Z_i\}_{i=1}^M$ are independent, bounded random variables with
$
\mathbb{E}[Z_i]=\mu,
\mathrm{Var}(Z_i)=\sigma^2.
$
Then, for large $M$, by the central limit theorem, the margin admits the Gaussian approximation
\[
\Delta
\approx
\mathcal{N}(M\mu, M\sigma^2),
\]
and the prediction accuracy is approximated by
\[ \Pr(\hat y=y^*) \approx \Phi\!\left(\sqrt{M}\,\mu/\sigma\right). \]
where $\Phi(\cdot)$ denotes the standard normal cumulative distribution function.
\end{proposition}

\begin{proof}
A complete proof is provided in Appendix~\ref{app:proof-margin-accuracy}.
\end{proof}






Proposition~\ref{thm:margin-accuracy} shows that accuracy is determined by the signal-to-noise ratio (SNR) $\sqrt{M}\mu/\sigma$, where $\mu$ is the expected per-path advantage of correct answer and $\sqrt{M}$ reflects the amplification of aggregation. Assuming $\sigma$ approximately fixed for simplicity, accuracy improves monotonically with $M$ and $\mu$. Crucially, the direction of amplification is governed by the sign of $\mu$: when $\mu > 0$, aggregation reinforces the correct answer; when $\mu < 0$, it amplifies the strongest incorrect answer instead. 

\begin{figure}[t]
    \centering

    \begin{subfigure}[t]{0.24\linewidth}
        \centering
        \includegraphics[width=\linewidth]{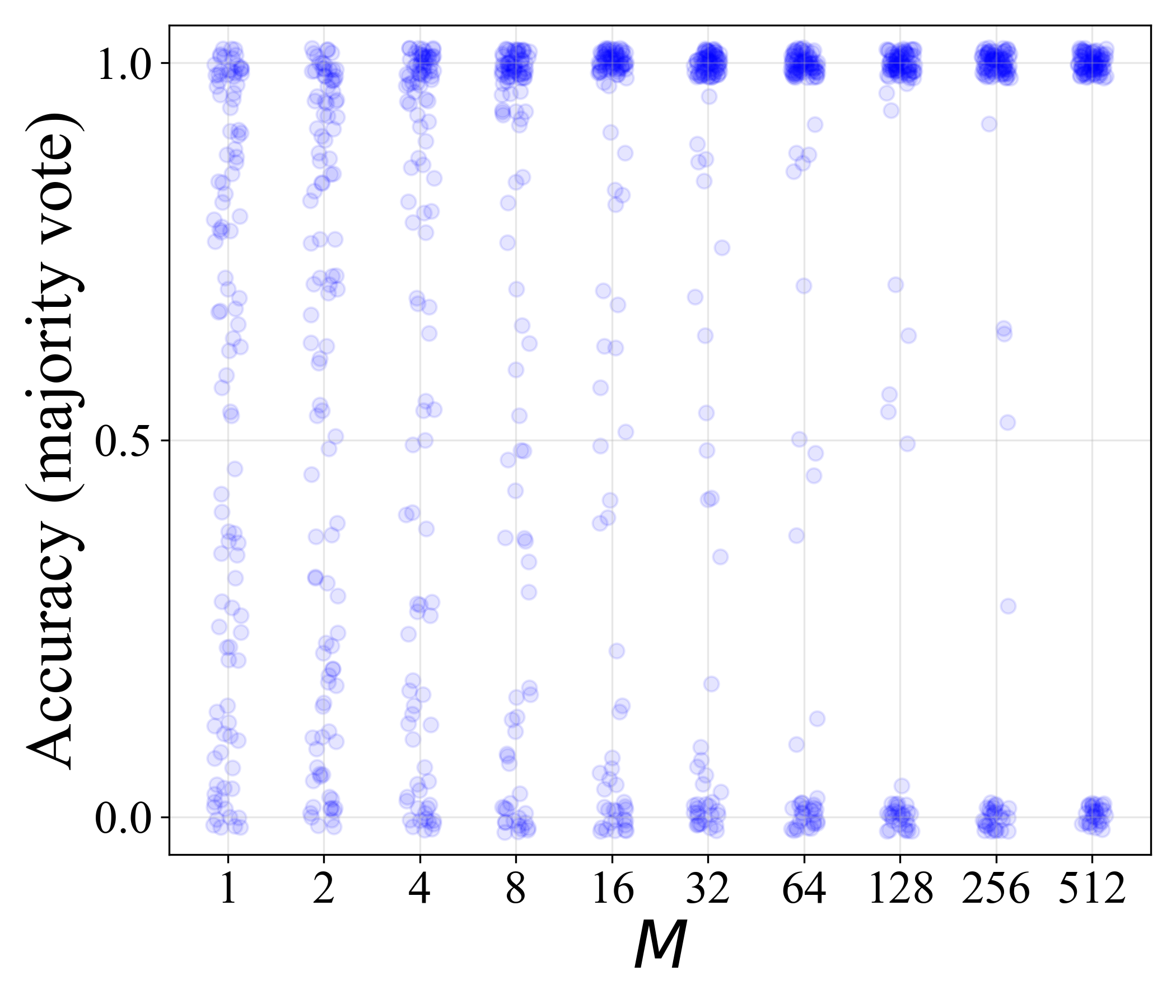}
        \vspace{-0.15in}
        \caption{SC}
        \label{fig:acc_vs_N_jitter}
    \end{subfigure}
    \hfill
    \begin{subfigure}[t]{0.24\linewidth}
        \centering
        \includegraphics[width=\linewidth]{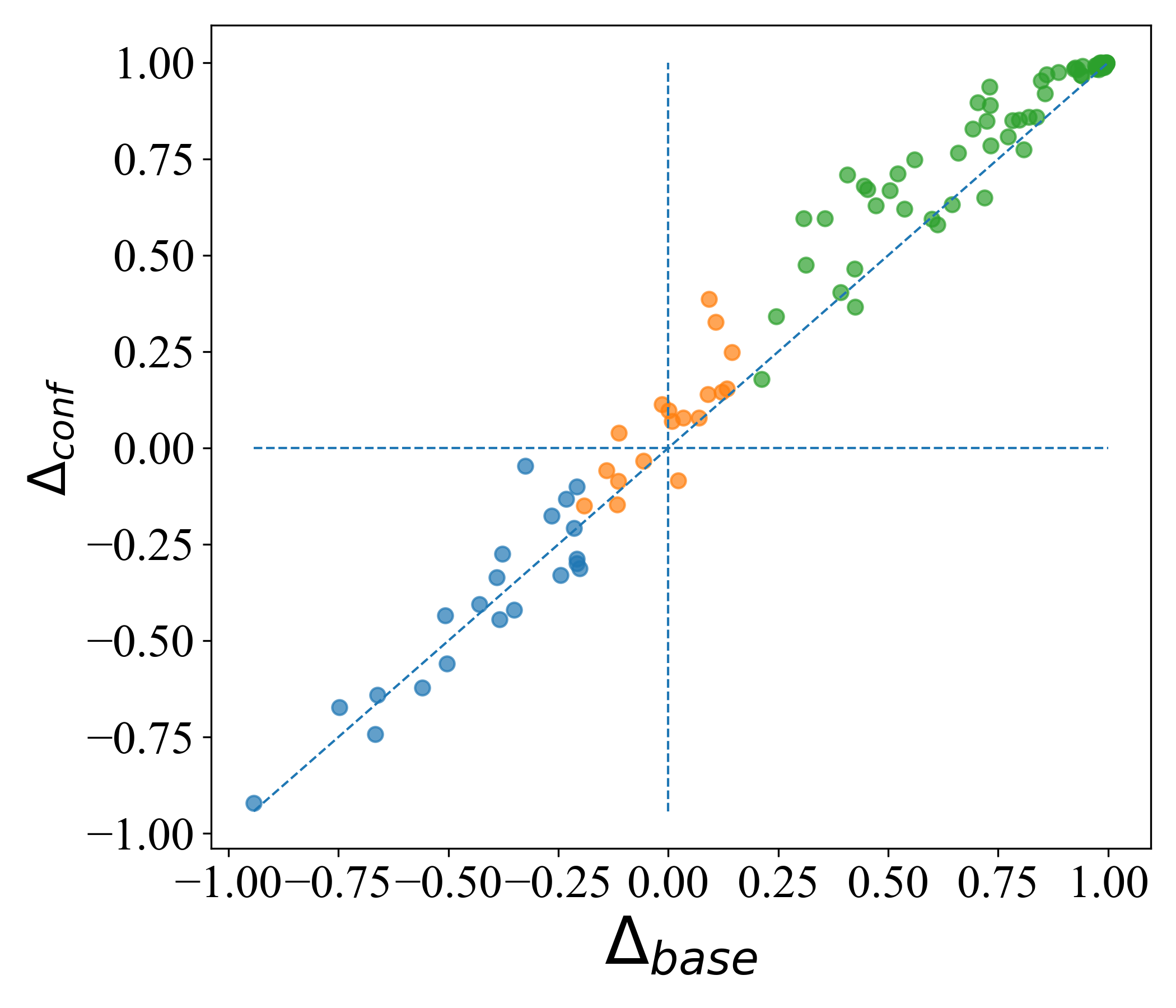}
        \vspace{-0.15in}
        \caption{SCert-Borda}
        \label{fig:margin_margin}
    \end{subfigure}
    \hfill
    \begin{subfigure}[t]{0.24\linewidth}
        \centering
        \includegraphics[width=\linewidth]{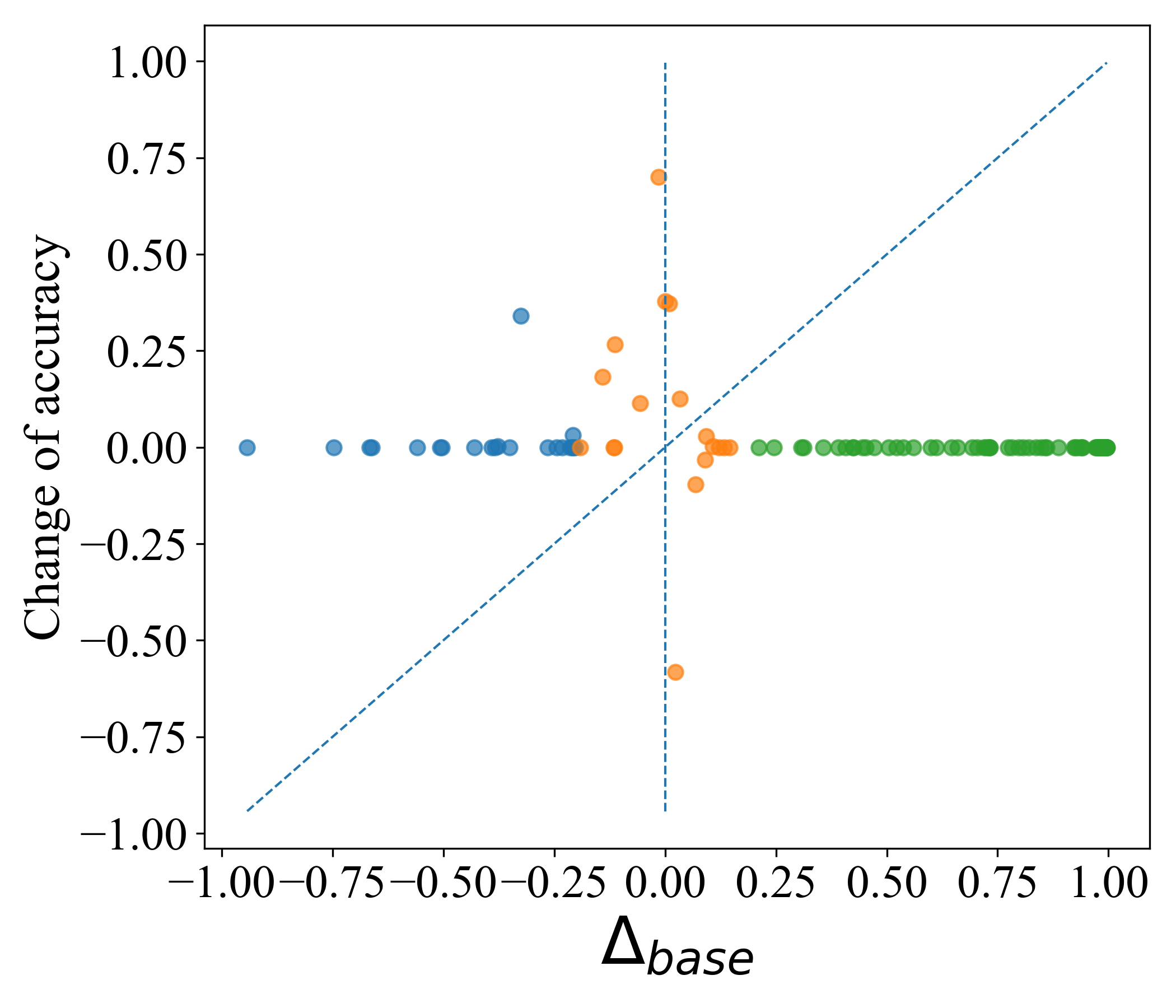}
        \vspace{-0.15in}
        \caption{SCert-Borda}
        \label{fig:margin_acc_end}
    \end{subfigure}
    \hfill
    \begin{subfigure}[t]{0.24\linewidth}
        \centering
        \includegraphics[width=\linewidth]{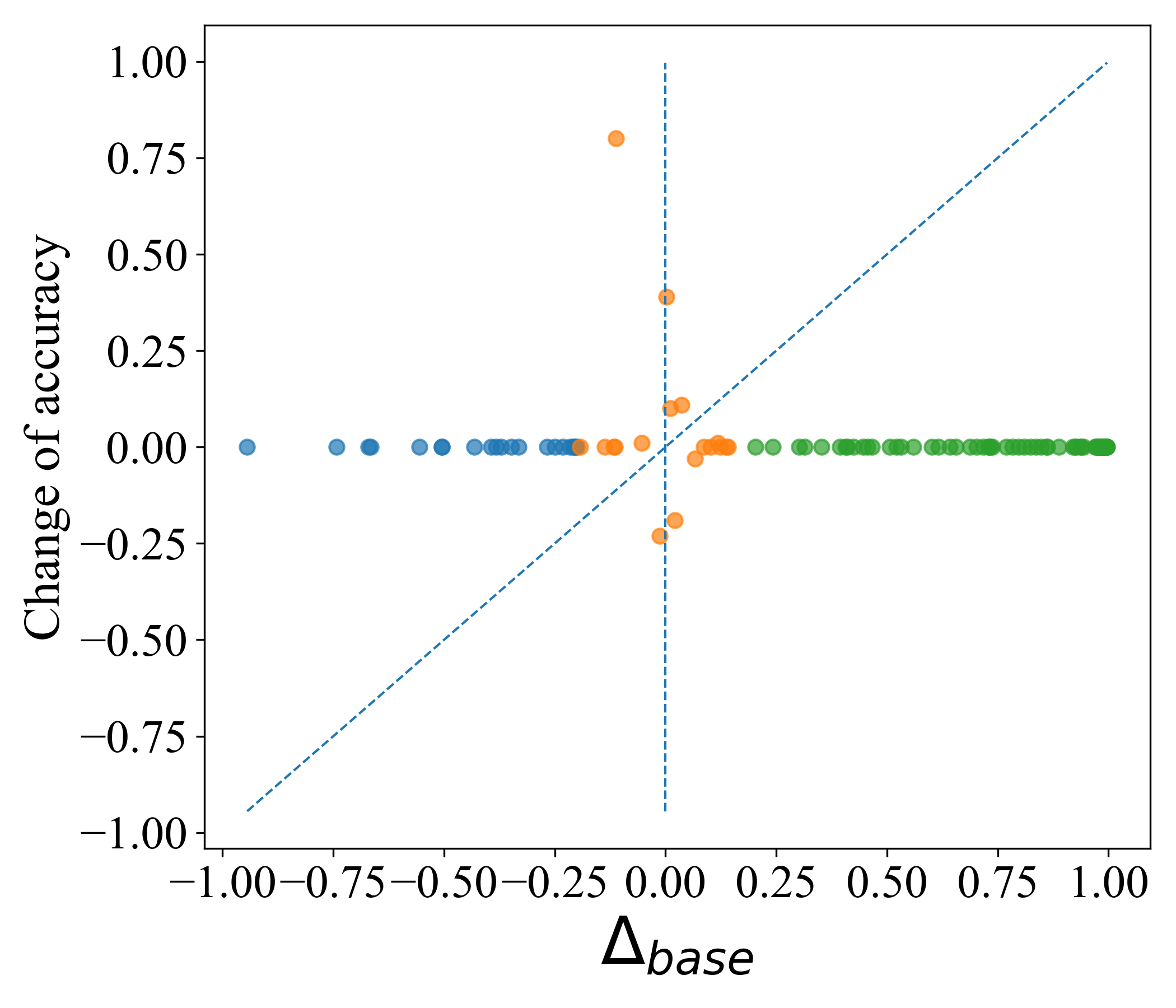}
        \vspace{-0.15in}
        \caption{SCert-Borda (prefix)}
        \label{fig:margin_acc_begin}
    \end{subfigure}

    \caption{Studies on the interaction between confidence scores and aggregation using Qwen3-8B~\cite{yang2025qwen3} (a) shows problem accuracy distribution of SC as $M$ increases. The point is jittered to show density.
    Prefix length in (d) is 4096 which is 22\% of full path length in (b)(c). (b)(c)(d) fixes $M=200$.
    }
    \label{fig:score_curve}
\end{figure}

To verify this, we run experiments using SC in Figure~\ref{fig:acc_vs_N_jitter}. Each point represents the estimated accuracy of a single problem under $M$ aggregated paths, where accuracy is estimated by repeating the aggregation 100 times. As $M$ increases, per-problem accuracies polarize toward either $1.0$ or $0$, consistent with the sign-driven amplification predicted by the theorem.

\subsection{Confidence Scores Are Most Effective Near the Decision Boundary}

As shown above, accuracy is governed by $\mu$ and $M$. Obviously, $\mu$ varies across problems: for example, easier problems tend to have positive $\mu$, while harder ones can have negative $\mu$. We denote $\mu$ under SC as $\mu_{\text{base}}$. After using confidence scores $\mu_{\text{base}}$ is shifted to $\mu_{\text{conf}}$. In this subsection, we examine how confidence scores influence $\mu$ and for how this shift influences TTS.

\textbf{Confidence scores are most effective near the decision boundary.} Figure~\ref{fig:margin_margin} plots $\mu_{\text{conf}}$ against $\mu_{\text{base}}$ for individual problems. For problems with positive or near-zero $\mu_{\text{base}}$ (green and orange points), confidence scores consistently increase $\mu$, with some near-zero points even flipping sign. For strongly negative $\mu_{\text{base}} < -0.2$ (blue), the shift is noisy with no clear trend.

Nevertheless, whether $\mu$ improves or worsens in this region has little consequence for accuracy. As shown in Figure~\ref{fig:margin_acc_end}, which reports the change in accuracy after introducing confidence scores across different values, problems with $\mu_{\text{base}} < -0.2$ or $\mu_{\text{base}} > 0.2$ show almost no change in accuracy. This is because when $|\mu_{\text{base}}|$ is large, aggregation over even a moderate $M$ is sufficient to converge to the dominant answer, and the outcome becomes insensitive to shifts in $\mu$.

In contrast, for problems near the decision boundary (i.e., $\mu_{\text{base}}$ is around 0), the shift in $\mu$ is important in two ways. \underline{First}, it enlarges the small positive margins, making the correct answer more stable under the noise in sampling process. \underline{Second}, it can flip the sign of $\mu$, reversing the direction of amplification from incorrect to correct. These boundary problems are exactly where confidence scores provide the greatest benefit. As confirmed in Figure~\ref{fig:margin_acc_end}, accuracy improvements after introducing confidence scores are concentrated among problems with $\mu_{\text{base}} \approx 0$. More importantly, even the shift is small, the aggregation over $M$ paths can amplify it.



\textbf{Prefix signals are sufficient near the boundary.}
Since small shifts in $\mu$ can be disproportionately amplified by aggregation, a natural question is whether prefix-based confidence scores can provide sufficient signal. Prefixes carry less information than full paths (see Appendix~\ref{appd:prefix_score} for a quantitative analysis). Nevertheless, the experiment in  Figure~\ref{fig:margin_acc_begin} shows that they can achieve competitive influence to accuracy after aggregation. Using the same experimental setup as Figure~\ref{fig:margin_acc_end}, we replace the full-path confidence score with prefix score. Full-path score improves accuracy by 1.71\% on average, while prefix score achieves 0.91\%. The prefix recovers \textbf{53\%} of the accuracy gain from full-path scoring, but only uses \textbf{22\%} of the tokens. This suggests that prefix signals are not only actionable but also substantially more token-efficient than full-path generation. This motivates the token reallocation framework we introduce in the following Section~\ref{sec:framework}.

\section{Prefix-Guided Sampling}
\label{sec:framework}

Building on the above analysis, we propose a principled framework for parallel test-time scaling that reallocates tokens toward prefix exploration.

\subsection{Method Framework}

\begin{wrapfigure}{r}{0.4\linewidth}
    \vspace{-13pt}
    \centering
    \includegraphics[width=\linewidth]{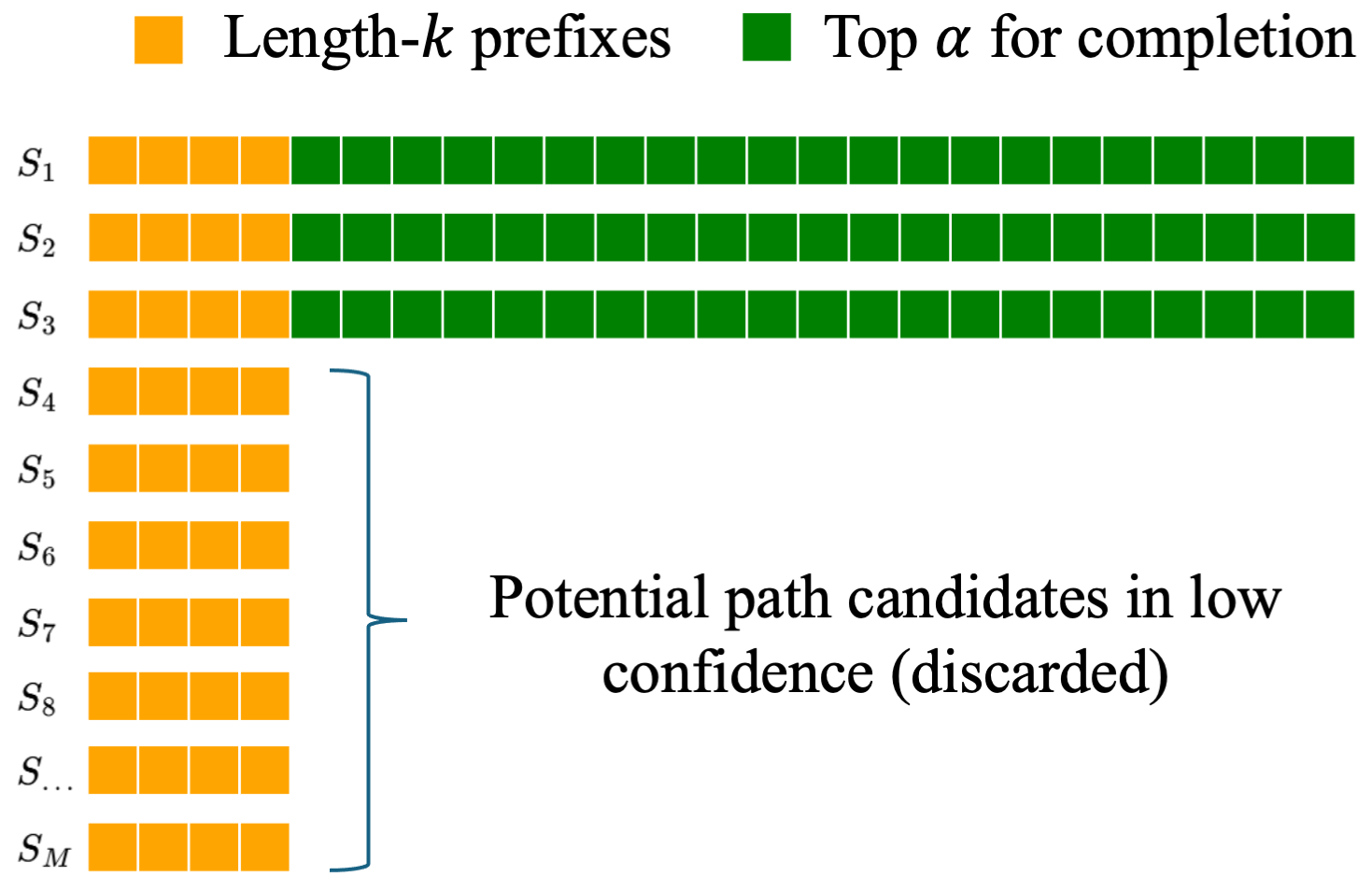}
    \vspace{-15pt}
    \caption{PreG framework}
    \label{fig:framework}
    \vspace{-15pt}
\end{wrapfigure}

We propose \textbf{Prefix-Guided Sampling} (PreG), a simple yet effective framework that is both model-agnostic and score-agnostic, consistently improving performance across aggregation strategies. As shown in Figure~\ref{fig:framework}, PreG operates in two stages: prefix generation and path completion.

\textbf{Stage 1: Prefix generation.} We first generate $M_{\text{prefix}}$ prefixes in parallel, each of length $k$. For each prefix, we compute an internal confidence score. (All the definitions of the confidence score we tested in this work can be found in Appendix~\ref{appd:score_def}. We additionally introduce a new confidence score, \textit{Conf-Lens}, derived from the logit lens~\cite{logitlens2020}.) The prefixes are then ranked by confidence scores to identify the most potential candidates.

\textbf{Stage 2: Path selection and completion.} We select the top fraction $\alpha$ of prefixes and continue generation to complete paths of length $T$, yielding $M_{\text{comp}} = \alpha M_{\text{prefix}}$ complete paths. Under the token budget $B$, the prefix stage consumes $M_{\text{prefix}} \cdot k$ tokens and the completion stage consumes $M_{\text{comp}} \cdot (T - k)$ tokens, giving
\[
M_{\text{comp}} = \frac{\alpha B}{k + \alpha(T - k)}.
\]
These complete paths are then aggregated via
\[
\hat{y} = \arg\max_y \sum_{i=1}^{\alpha M_\text{prefix}} W(S_i^{\text{comp}}) \cdot \mathbf{1}[y_i = y],
\]
where $W(\cdot)$ is determined by the confidence score of the completed path $S_i^{\text{comp}}$ and strategies such as majority voting, weighted voting, Borda voting~\cite{kangscalable}, or top-$10\%$ voting~\cite{fu2025deep, zhang2025tokur}.

\paragraph{Why PreG works.} Compared to vanilla parallel TTS, PreG improves $\mu$ in two ways: it filters out low-quality prefixes, and it explores a larger pool of $M_{\text{prefix}} > M_{\text{base}} =  B/T$ candidates before selecting the best $\alpha$ fraction. The cost is fewer completed paths. However, this is naturally mitigated by the aggregation mechanism. When $|\mu_{\text{base}}|$ is large, aggregation already converges to the dominant answer quickly, so the reduction in $M$ has little effect on accuracy. When $\mu_{\text{base}} \approx 0$, the improvement in $\mu$ is amplified, sometimes even flipping the direction of amplification.

\textbf{System efficiency.} Beyond algorithmic gains, PreG also offers practical advantages at the system level: all prefixes are generated in parallel to a fixed length, requiring no dynamic stopping conditions, cross-path comparisons, or prefill operations, making it better suited for parallel system design.

\textbf{Finite-sample regime.}
When $M$ is small, the signal-to-noise ratio $\sqrt{M}\mu/\sigma$ can be close to zero even when $\mu > 0$, meaning aggregation is dominated by sampling noise. In this regime, reducing $M$ through filtering can cause a large accuracy drop unless the gain in $\mu$ is sufficiently large. 
Empirically, however, our method already outperforms the vanilla baseline at values of $M$ well below those typical in standard parallel sampling setups.





\subsection{Analysis on $k$ and $\alpha$}
\label{sec:hyper}

Proposition~\ref{thm:margin-accuracy} shows that accuracy is governed by the signal-to-noise ratio (SNR) $S = \sqrt{M_{\text{comp}}}\mu/\sigma$. Under a fixed budget $B$, both $M$ and $\mu$ depend on the hyperparameters $\alpha$ and $k$, creating a trade-off: increasing $\alpha$ yields more complete paths but weaker filtering, while increasing $k$ improves signal quality but reduces $M_{\text{comp}}$. The following propositions characterize the optimal choices.

\paragraph{Optimal prefix length $k$.} 
Assume $\mu(k) = \mu_0 + \beta k$ grows linearly with prefix length. Then $V(k) = \sqrt{M(k)}\,\mu(k)$ has no interior maximum over $k \in [k_{\min}, k_{\max}]$, and the optimal prefix length satisfies $k^* \in \{k_{\min}, k_{\max}\}$.
A complete derivation is provided in Appendix~\ref{app:proof-boundary-k}.

For simplicity, we assumed $\mu(k) = \mu_0 + \beta k$ grows linearly with prefix length. However, as shown in Appendix~\ref{appd:prefix_score}, the signal grows very slowly and is near zero for the first 1000--2000 tokens, suggesting that $k_{\min}$ should not be set too small. In this work, the smallest $k$ we evaluate is 4000 tokens.

\paragraph{Optimal filtering ratio $\alpha$.} Under a fixed budget $B$, increasing $\alpha$ yields more complete paths but admits more low-quality ones, while decreasing $\alpha$ improves path quality but reduces $M_{\text{comp}}$. Taking $\alpha \to 0$ collapses the SNR regardless of $\mu$; taking $\alpha = 1$ reduces to the vanilla method. Unless the vanilla method outperforms PreG, the optimal $\alpha^*$ lies between the two extremes.

These theoretical findings are consistent with our empirical results in Section~\ref{sec:hyperpara}.
\section{Experiments}
\label{sec:exp}

In this section, we evaluate PreG across multiple reasoning benchmarks, confidence scores and aggregation strategies. We further ablate the finite-sample regime, analyze the sensitivity to the two key hyperparameters, and evaluate system-level efficiency. We first present the experimental settings.

\textbf{LLMs and Datasets.} We evaluate two reasoning models: Qwen3-8B (Qwen)~\cite{yang2025qwen3} and Llama-3.1-Nemotron-Nano-4B-v1.1 (Llama)~\cite{bercovich2025llama}. We benchmark on four math competition datasets~\cite{dekoninck2026matharena}, AIME25, BRUMO25, HMMT25-Feb, and CMIMC25, as well as GPQA-Diamond~\cite{rein2023gpqa}, a dataset spanning biology, physics, and chemistry.

\textbf{Baselines.} We consider vanilla parallel TTS methods, early-stopping methods and PRM. The first category includes Self-Consistency and six methods based on six token-level confidence scores, likelihood (LL)~\cite{taubenfeld2025confidence}, entropy~\cite{sharma2025sequential}, Gini Impurity~\cite{loh2011classification, kangscalable}, Diff-token, Conf-Lens, DeepConf (DC)~\cite{fu2025deep}, and Self-Certainty (SCert)~\cite{kangscalable}. They are combined with four aggregation strategies, majority voting, weighted voting, Borda voting~\cite{kangscalable} and top-10\% voting (W-10\%)~\cite{fu2025deep, zhang2025tokur}. Early-stopping methods include ESC~\cite{liescape}, ST-BoN~\cite{wang2025sampling}, and CGES~\cite{aghazadeh2025cges}. The PRM method uses weighted voting. The PRMs are Qwen2.5-Math-PRM-7B~\cite{prmlessons} for math, and Skywork-Reward-V2-Qwen3-8B~\cite{liu2025skywork} for GPQA-D.

\textbf{Implementation and evaluation.} We pre-generate 512 paths per problem at temperature $= 1.1$ on H200s. For each evaluation, we sample from these paths without replacement over 100 random seeds to simulate the TTS process. Confidence scores in vanilla methods are computed over the last 2048 tokens of each path; prefix scores in PreG use the first 4096 tokens (${\sim}22\%$ of Qwen output and ${\sim}18\%$ of Llama output). We set $\alpha=0.3$ for all PreG configurations. 
In this setting, we have $M_{\text{prefix}} = 2.2M_{\text{base}}$ and $M_{\text{comp}} = 0.66M_{\text{base}}$ for Qwen, and $M_{\text{prefix}} = 2.31M_{\text{base}}$ and $M_{\text{comp}} = 0.69M_{\text{base}}$ for Llama.
Performance of a parallel TTS method is reported as accuracy averaged over 100 attempts. Effectiveness of a confidence score is measured by AUROC.

\begin{table}
  \caption{Accuracy under fixed budgets (100k, 300k, 1000k tokens). Results exceeding and falling below the corresponding vanilla baseline are underlined in \ulgood{green} and \ulbad{red}, respectively. The best result per dataset-budget combination is shown in \textbf{bold}.} 
  \label{tab:main_results}
  \centering
  \resizebox{\linewidth}{!}{
  \begin{tabular}{clccccc}
    \toprule
    \multicolumn{2}{c}{\multirow{2}{*}{QWen3-8B}} & AIME25 & BRUMO & HMMT & CMIMC & GPQA-D \\
     & & (100k/300k/1000k) & (100k/300k/1000k) & (100k/300k/1000k) & (100k/300k/1000k) & (100k/300k/1000k) \\

    \midrule

    \multicolumn{2}{c}{Pass@1} & 67.94 & 69.64 & 45.30 & 46.14 & 58.90 \\


    \multicolumn{2}{c}{Self-Consistency} & 76.53 / 78.43 / 79.40 & 76.93 / 78.70 / 79.90 & 52.37 / 56.10 / 57.33 & 52.78 / 55.50 / 56.33 & 68.90 / 69.87 / 70.75 \\ 
    \multicolumn{2}{c}{PRM} & 78.17 / 78.43 / 78.97 & 78.17 / 78.93 / 80.27 & 56.60 / 57.00 / 58.10 & 55.17 / 55.53 / 56.55 & 67.88 / 68.43 / 69.41 \\

    \multicolumn{2}{c}{ESC} & \textbf{78.33} / 78.77 / \textbf{81.00} & 77.87 / 79.47 / 80.73 & 55.87 / 57.13 / 58.03 & 55.10 / 55.85 / 56.23 & 69.24 / 70.17 / ~NA~~ \\
    \multicolumn{2}{c}{ST-BoN} & 75.40 / 78.70 / 80.60 & 75.13 / 78.00 / 79.73 & 49.40 / 53.73 / 55.13 & 49.48 / 54.65 / 56.13 & 67.77 / 67.34 / 68.92 \\
    \multicolumn{2}{c}{CGES} & 70.43 / 75.10 / ~NA~~ & 75.20 / 78.17 / ~NA~~ & 51.40 / 53.13 / ~NA~~ & 52.18 / 54.10 / ~NA~~ & ~~NA~ / ~~NA~~ / ~NA~~ \\

    \midrule

    \multirow{6}{*}{Borda} & LL & 75.30 / 77.87 / 79.33 & 75.87 / 78.87 / 79.27 & 50.93 / 55.73 / 56.37 & 50.25 / 55.30 / 57.65 & 69.61 / 70.12 / 70.81 \\
    & LL-PreG & \ulbad{74.90} / \ulbad{77.43} / \ulbad{79.30} & \ulgood{78.43} / \ulgood{81.43} / \ulgood{82.77} & \ulgood{53.27} / \ulgood{57.80} / \ulgood{60.00} & \ulgood{51.42} / \ulgood{56.95} / \ulgood{58.43} & \ulbad{69.07} / \ulbad{69.39} / \ulbad{69.67} \\
    & DC & 76.27 / 78.43 / 78.83 & 77.43 / 80.33 / 81.63 & 55.37 / 59.00 / 60.47 & 53.65 / 58.87 / 60.72 & 69.58 / 70.12 / 70.58 \\
    & DC-PreG & \ulbad{76.20} / \ulgood{\textbf{79.07}} / \ulgood{79.93} & \ulgood{79.77} / \ulgood{82.07} / \ulgood{83.03} & \ulgood{56.63} / \ulgood{59.90} / \ulgood{61.47} & \ulgood{\textbf{55.45}} / \ulgood{60.68} / \ulgood{63.35} & \ulgood{69.63} / \ulbad{69.95} / ~NA~~ \\
    & SCert & 76.33 / 78.33 / 78.70 & 77.77 / 80.93 / 82.33 & 56.10 / 58.87 / 60.03 & 54.47 / 58.73 / 60.15 & 69.46 / 69.81 / 70.32 \\
    & SCert-PreG & \ulgood{76.40} / \ulgood{78.67} / \ulgood{80.60} & \ulgood{\textbf{\textbf{80.73}}} / \ulgood{\textbf{\textbf{82.70}}} / \ulgood{83.40} & \ulgood{\textbf{\textbf{57.10}}} / \ulgood{59.00} / \ulgood{60.20} & \ulgood{55.20} / \ulgood{\textbf{\textbf{60.78}}} / \ulgood{62.55} & \ulgood{\textbf{69.62}} / \ulgood{\textbf{70.05}} / \ulgood{\textbf{\textbf{70.45}}} \\

    \midrule

    \multirow{6}{*}{W-10\%} & LL & 67.83 / 65.60 / 73.10 & 70.60 / 69.60 / 75.43 & 46.47 / 46.37 / 51.43 & 46.00 / 44.62 / 48.98 & 66.63 / 67.89 / 68.36 \\
    & LL-PreG & \ulbad{67.03} / \ulgood{71.30} / \ulgood{76.27} & \ulgood{72.57} / \ulgood{76.97} / \ulgood{81.23} & \ulgood{48.50} / \ulgood{52.40} / \ulgood{57.60} & \ulgood{46.52} / \ulgood{47.80} / \ulgood{55.20} & \ulgood{66.79} / \ulgood{68.41} / \ulgood{68.71} \\
    & DC & 72.70 / 71.90 / 76.50 & 74.87 / 75.70 / 80.93 & 53.53 / 55.60 / 60.23 & 52.68 / 54.43 / 58.95 & 67.27 / 68.51 / 69.49 \\
    & DC-PreG & \ulbad{71.63} / \ulgood{75.70} / \ulgood{78.57} & \ulgood{76.73} / \ulgood{80.20} / \ulgood{82.07} & \ulbad{52.93} / \ulgood{58.73} / \ulgood{61.37} & \ulbad{51.18} / \ulgood{56.73} / \ulgood{64.72} & \ulbad{67.03} / \ulgood{69.03} / \ulgood{69.83} \\
    & SCert & 72.37 / 72.53 / 76.80 & 76.00 / 77.47 / 80.77 & 54.53 / 57.53 / 60.93 & 54.30 / 55.33 / 59.55 & 67.57 / 68.44 / 69.28 \\
    & SCert-PreG & \ulgood{72.40} / \ulgood{75.73} / \ulgood{78.70} & \ulgood{77.80} / \ulgood{81.50} / \ulgood{\textbf{\textbf{83.47}}} & \ulbad{53.50} / \ulgood{\textbf{\textbf{60.53}}} / \ulgood{\textbf{\textbf{62.67}}} & \ulbad{52.90} / \ulgood{58.77} / \ulgood{\textbf{\textbf{66.17}}} & \ulbad{67.18} / \ulgood{68.92} / \ulgood{69.61} \\

    \midrule
    \multicolumn{2}{c}{\multirow{2}{*}{Llama-Nemotron-4B}} & AIME25 & BRUMO & HMMT & CMIMC & GPQA-D \\
     & & (100k/300k/1000k) & (100k/300k/1000k) & (100k/300k/1000k) & (100k/300k/1000k) & (100k/300k/1000k) \\

    \midrule

    \multicolumn{2}{c}{Pass@1} & 56.41 & 63.76 & 41.11 & 37.88 & 52.51 \\

    \multicolumn{2}{c}{Self-Consistency} & 65.73 / 68.93 / 70.10 & 66.60 / 66.40 / 65.90 & 44.53 / 47.87 / 48.23 & 42.10 / 48.10 / 52.35 & 59.81 / 60.34 / 59.78 \\
    \multicolumn{2}{c}{PRM} & 67.63 / 69.30 / 69.97 & 67.20 / 66.50 / 65.97 & 47.77 / 48.30 / 48.50 & 46.40 / 47.75 / 51.15 & {60.82} / 61.43 / 59.36 \\
    \multicolumn{2}{c}{ESC} & 67.60 / 68.83 / 70.53 & 67.33 / 66.60 / 65.03 & 47.27 / 47.50 / 48.20 & \textbf{46.72} / 47.98 / 53.10 & 58.18 / 58.70 / 59.63 \\
    \multicolumn{2}{c}{ST-BoN} & 63.80 / 68.03 / 68.80 & 66.53 / 66.77 / 65.50 & 43.80 / 47.30 / 46.77 & 40.65 / 45.45 / 49.53 & 58.55 / 58.97 / 59.54 \\
    \multicolumn{2}{c}{CGES} & 61.00 / 65.10 / ~NA~~ & 63.73 / 62.47 / ~NA~~ & 42.87 / 45.00 / ~NA~~ & 42.95 / 44.27 / ~NA~~ & ~~NA~ / ~~NA~~ / ~NA~~ \\

    \midrule

    \multirow{6}{*}{Borda} & LL & 67.23 / 71.10 / 72.23 & 67.87 / 68.07 / 67.63 & 47.33 / 50.37 / 52.10 & 43.48 / 49.00 / 53.30 & 60.48 / 61.59 / 61.90 \\
    & LL-PreG & \ulgood{67.97} / \ulgood{71.13} / \ulgood{73.50} & \ulgood{68.47} / \ulgood{\textbf{\textbf{69.47}}} / \ulgood{70.30} & \ulgood{47.73} / \ulgood{\textbf{\textbf{50.93}}} / \ulgood{53.37} & \ulgood{43.60} / \ulbad{48.60} / \ulbad{53.02} & \ulgood{\textbf{61.43}} / \ulgood{\textbf{\textbf{62.56}}} / \ulgood{62.50} \\
    & DC & 67.50 / \textbf{\textbf{71.83}} / 73.13 & 67.60 / 68.00 / 67.47 & 47.87 / 50.67 / 52.03 & 43.80 / 48.98 / 53.57 & 60.39 / 61.47 / 61.77 \\
    & DC-PreG & \ulgood{68.20} / \ulbad{71.47} / \ulgood{74.63} & \ulgood{68.00} / \ulgood{69.10} / \ulgood{69.23} & \ulgood{\textbf{\textbf{48.57}}} / \ulbad{50.60} / \ulgood{53.13} & \ulgood{44.82} / \ulgood{\textbf{49.75}} / \ulgood{\textbf{\textbf{54.15}}} & \ulgood{61.22} / \ulgood{62.18} / \ulgood{\textbf{62.23}} \\
    & SCert & 67.73 / 71.73 / 73.30 & 67.73 / 67.93 / 67.33 & 47.47 / 50.70 / 51.63 & 43.40 / 49.15 / 53.50 & 60.21 / 61.11 / 61.27 \\
    & SCert-PreG & \ulgood{\textbf{68.60}} / \ulbad{71.70} / \ulgood{\textbf{\textbf{74.93}}} & \ulgood{\textbf{\textbf{68.57}}} / \ulgood{68.80} / \ulgood{68.17} & \ulgood{48.50} / \ulgood{50.77} / \ulgood{52.50} & \ulgood{44.90} / \ulgood{49.55} / \ulgood{53.53} & \ulgood{60.66} / \ulgood{61.57} / \ulgood{61.47} \\

    \midrule
    
    \multirow{6}{*}{W-10\%} & LL & 63.87 / 64.37 / 70.80 & 66.40 / 67.07 / 70.67 & 45.80 / 47.90 / 50.17 & 42.53 / 43.40 / 48.90 & 59.83 / 60.27 / 62.89 \\
    & LL-PreG & \ulbad{62.53} / \ulgood{67.63} / \ulgood{74.73} & \ulgood{67.10} / \ulgood{68.57} / \ulgood{\textbf{\textbf{71.50}}} & \ulbad{45.17} / \ulgood{49.50} / \ulgood{\textbf{\textbf{53.63}}} & \ulbad{41.68} / \ulgood{43.87} / \ulgood{49.50} & \ulgood{59.93} / \ulgood{62.21} / \ulgood{63.50} \\
    & DC & 64.83 / 66.33 / 72.27 & 67.20 / 67.27 / 70.53 & 47.20 / 49.33 / 51.70 & 43.10 / 44.65 / 50.75 & 60.05 / 60.71 / 63.15 \\
    & DC-PreG & \ulbad{63.87} / \ulgood{69.07} / \ulgood{73.30} & \ulgood{67.20} / \ulgood{68.60} / \ulgood{71.13} & \ulbad{46.23} / \ulgood{50.90} / \ulgood{53.20} & \ulbad{42.93} / \ulgood{46.57} / \ulgood{52.38} & \ulbad{59.86} / \ulgood{62.09} / \ulgood{63.26} \\
    & SCert & 65.47 / 67.23 / 71.83 & 67.20 / 67.47 / 70.17 & 46.90 / 49.97 / 52.93 & 42.63 / 44.40 / 51.50 & 59.77 / 60.87 / 62.49 \\
    & SCert-PreG & \ulbad{63.90} / \ulgood{69.20} / \ulgood{73.67} & \ulbad{66.57} / \ulgood{69.30} / \ulgood{70.33} & \ulbad{46.60} / \ulgood{49.97} / \ulbad{52.87} & \ulbad{42.00} / \ulgood{46.05} / \ulgood{51.80} & \ulbad{59.69} / \ulgood{61.91} / \ulbad{62.15} \\

    \bottomrule
  \end{tabular}
  }
\end{table}

\subsection{Main Results}
\label{sec:main_results}

We first compare methods under a fixed token budget, evaluating at three budget levels, 100k, 300k, and 1000k tokens, where the corresponding $M_{\text{base}}$ is approximately 5.4, 16.2, and 54.1 for Qwen and 4.6, 13.8, and 46.1 for Llama. Due to space constraints, the main results include three representative confidence scores (LL, DC, and SCert) and the two best-performing aggregation strategies among the baselines: Borda voting and W-10\%. Borda voting assigns weights based on score rankings; W-10\% retains the top 10\% of $M_{\text{base}}$ paths and applies weighted voting. {PreG} uses the same aggregation strategies. For W-10\% in PreG, this results in two rounds of path filtering: the first uses the prefix score to retain $M_{\text{comp}}$ paths, and the second uses the full score to retain the 10\% of $M_{\text{prefix}}$ for weighted voting. All methods are compared under the same total token budget. We report the results in Table~\ref{tab:main_results}.

\textbf{Overall}, our method achieves strong and consistent improvements across datasets and models.
Specifically, (1) our approach attains the best performance among all compared methods on the majority
of datasets and model configurations, and (2) it yields stable and significant gains over the corresponding vanilla confidence-based
baseline.
We now discuss from the following perspectives.


\textbf{Performance Across Different Datasets.}
Our method improves performance on all datasets, with gains particularly stable when the token budget exceeds 300k (smaller budgets are analyzed in Section~\ref{sec:small_budget}).
(1) Among the 15 dataset–budget combinations, {PreG} achieves the best result in 13 cases under Qwen and 14 cases under LLaMA.
Improvements can be substantial: on CMIMC (1000k budget), {SCert-PreG} (W-10\%) outperforms all baselines by 5.4\% under Qwen, where SC itself gains only $\sim$10\% over Pass@1. (2) Across all settings, our method consistently outperforms the corresponding vanilla confidence methods by a large margin.

\textbf{Different Voting Methods.}
PreG improve both Borda and W-10\% but exhibit different behaviors. PreG with Borda is more stable across budgets, outperforming the vanilla confidence baselines in most configurations and yielding at least 2\% improvement on Qwen in most cases. PreG with W-10\% performs poorly under small budgets—roughly matching the baseline at 100k for Qwen and even degrading slightly for LLaMA. But it is more effective at 1,000k budgets, achieving the best result among all methods in 5 of 8 dataset–model math combinations.
We attribute this to the aggressive filtering of W-10\%: at 100k, only 1 path survives to vote, making results highly unstable. At large budgets, this aggressiveness becomes an advantage: if correct paths are the majority, discarding low-confidence paths does not hurt; if incorrect paths dominate, the filtering is more likely to remove them. Thus, at large budgets, W-10\% can amplify the signal from high-quality paths more effectively. We also report the results of other aggregation in Appendix~\ref{appd:other_vote_score}.

\textbf{Different Confidence Scores.}
{PreG} achieves better absolute performance when paired with stronger confidence scores, but yields more pronounced relative improvements on weaker scores. DC and SCert are the two strongest confidence scores, and the best overall results concentrate around these methods. For weaker scores (represented by LL), {PreG} consistently closes the gap with vanilla DC and SCert, often matching DC-{PreG} and SCert-{PreG}. The results of other scores are in Appendix~\ref{appd:other_vote_score}.

\textbf{Comparison with early stopping and PRM-based methods.}
{PreG} outperforms all three early stopping baselines, demonstrating more effective utilization of the token budget. Our method also surpasses the PRM baseline evaluated in our experiments.







\subsection{Token Efficiency}
\label{sec:exp_token_eff}

\begin{table*}[t]
\centering
\begin{minipage}{0.67\textwidth}
    \centering
    \caption{Token efficiency. Underline colors indicate \ulgood{$>$50\% token savings}, \ulmid{0--50\% token savings}, and \ulbad{no savings}.}
    \resizebox{\linewidth}{!}{
    \begin{tabular}{ccccccc}
    \toprule
    \multicolumn{1}{c}{Qwen3} & Score & AIME25 & BRUMO & HMMT & CMIMC & GPQA-D \\

    \midrule

    \multirow{3}{*}{Borda} & LL & \ulbad{144.25\%} & \ulgood{41.83\%} & \ulgood{46.64\%} & \ulmid{50.87\%} & \ulbad{NA} \\
    & DC & \ulmid{57.98\%} & \ulgood{20.50\%} & \ulmid{62.53\%} & \ulmid{54.90\%} & \ulbad{121.10\%} \\
    & SCert & \ulmid{55.07\%} & \ulgood{23.07\%} & \ulmid{63.08\%} & \ulgood{46.83\%} & \ulmid{61.76\%} \\
    \midrule
    \multirow{3}{*}{W-10\%} & LL & \ulgood{35.27\%} & \ulgood{33.87\%} & \ulgood{25.89\%} & \ulgood{33.64\%} & \ulmid{55.26\%} \\
    & DC & \ulgood{33.25\%} & \ulgood{44.09\%} & \ulgood{45.95\%} & \ulgood{47.92\%} & \ulmid{58.27\%} \\
    & SCert & \ulmid{54.18\%} & \ulgood{49.66\%} & \ulmid{51.29\%} & \ulgood{47.81\%} & \ulmid{62.54\%} \\

    \midrule
    \multicolumn{1}{c}{Llama} & Score & AIME25 & BRUMO & HMMT & CMIMC & GPQA-D \\
    
    \midrule
    
    \multirow{3}{*}{Borda} & LL & \ulgood{44.11\%} & \ulgood{29.36\%} & \ulmid{57.76\%} & \ulbad{114.34\%} & \ulgood{19.02\%} \\
    & DC & \ulmid{54.25\%} & \ulgood{29.23\%} & \ulmid{68.82\%} & \ulmid{84.91\%} & \ulgood{20.55\%} \\
    & SCert & \ulgood{42.16\%} & \ulgood{29.43\%} & \ulmid{53.62\%} & \ulmid{91.52\%} & \ulgood{26.02\%} \\
    \midrule
    \multirow{3}{*}{W-10\%} & LL & \ulgood{47.65\%} & \ulmid{59.69\%} & \ulgood{32.04\%} & \ulmid{62.70\%} & \ulmid{56.50\%} \\
    & DC & \ulmid{56.18\%} & \ulgood{40.28\%} & \ulmid{73.58\%} & \ulmid{67.53\%} & \ulmid{71.65\%} \\
    & SCert & \ulmid{54.33\%} & \ulgood{42.47\%} & \ulmid{91.36\%} & \ulmid{78.13\%} & \ulbad{NA} \\

    \bottomrule
  \end{tabular}
  \label{tab:fix_acc}
  }
\end{minipage}
\hfill
\begin{minipage}{0.305\textwidth}
    \centering
    \caption{Confidence signal effectiveness.}

\resizebox{\linewidth}{!}{
  \begin{tabular}{lcc}
    \toprule
    Qwen & Math & GPQA-D \\

    \midrule

    PRM & 0.6148 & 0.5122 \\

    LL & 0.5462 & 0.5660 \\
    LL-PreG & 0.5345 & 0.5055  \\
    DC & 0.7260 & 0.5651 \\
    DC-PreG & 0.5415 & 0.5481 \\
    SCert & 0.7409 & 0.5814 \\
    SCert-PreG & 0.5471 & 0.5359  \\

    \midrule

    Llama & Math & GPQA-D \\

    \midrule

    PRM & 0.5959 & 0.5532 \\
    
    LL & 0.6229 & 0.5670 \\
    LL-PreG & 0.5151 & 0.5394 \\
    DC & 0.6576 & 0.5785  \\
    DC-PreG & 0.5241 & 0.5360 \\
    SCert & 0.6624 & 0.5811  \\
    SCert-PreG & 0.5187 & 0.5278 \\

    \bottomrule
  \end{tabular}
  \label{tab:auroc}
  }

\end{minipage}
\end{table*}


In Table~\ref{tab:fix_acc}, we compare token efficiency against the vanilla confidence methods. We measure efficiency as the ratio of tokens required by PreG versus the corresponding vanilla method at a target accuracy. The target accuracy is defined as the accuracy achieved by the vanilla method at a 1,000k token budget. (If the vanilla method saturates before 1000k tokens, we use the actual token budget at which the target accuracy is first reached.)

Across all 60 configurations, {PreG} requires fewer tokens than the vanilla method in 54 cases, and in 27 of these requires less than 50\% of the tokens. Token savings are most pronounced on BRUMO, where up to $\sim$80\% of tokens can be saved. On mathematical datasets, Qwen saves over 40\% in all but one configuration (AIME, Borda, LL); LLaMA similarly achieves over 40\% savings in the majority of cases.
Overall, these results demonstrate PreG's effectiveness as a token-efficient inference strategy.



\subsection{Confidence Signal Effectiveness}
\label{sec:auroc}

Table~\ref{tab:auroc} compares the AUROC of different confidence scores, where ``Math'' denotes averaged results across four mathematical datasets. AUROC is first computed within each problem. The results are averaged over problems that are neither fully correct nor fully incorrect.


Comparing full generation versus prefix, prefix confidence is substantially weaker: SCert achieves an AUROC of 0.7409 on full-length paths but only 0.5471 on prefixes. Combined with the strong performance gains reported earlier, this supports our analysis that parallel TTS amplifies the signal near the decision boundary—demonstrating that even weak prefix signals can be effectively leveraged.

On GPQA-Diamond, AUROC scores are smaller than math data, which explains why adding confidence signals sometimes degrades performance. This confirms a key limitation of our method: its effectiveness is strongly tied to the quality of the underlying confidence signal. However, when that signal is reliable, our method substantially improves both performance and token efficiency.








    


\subsection{Hyperparameters}
\label{sec:hyperpara}

\begin{figure}[t]
    \centering

    \begin{subfigure}[t]{0.245\linewidth}
        \centering
        \includegraphics[width=\linewidth]{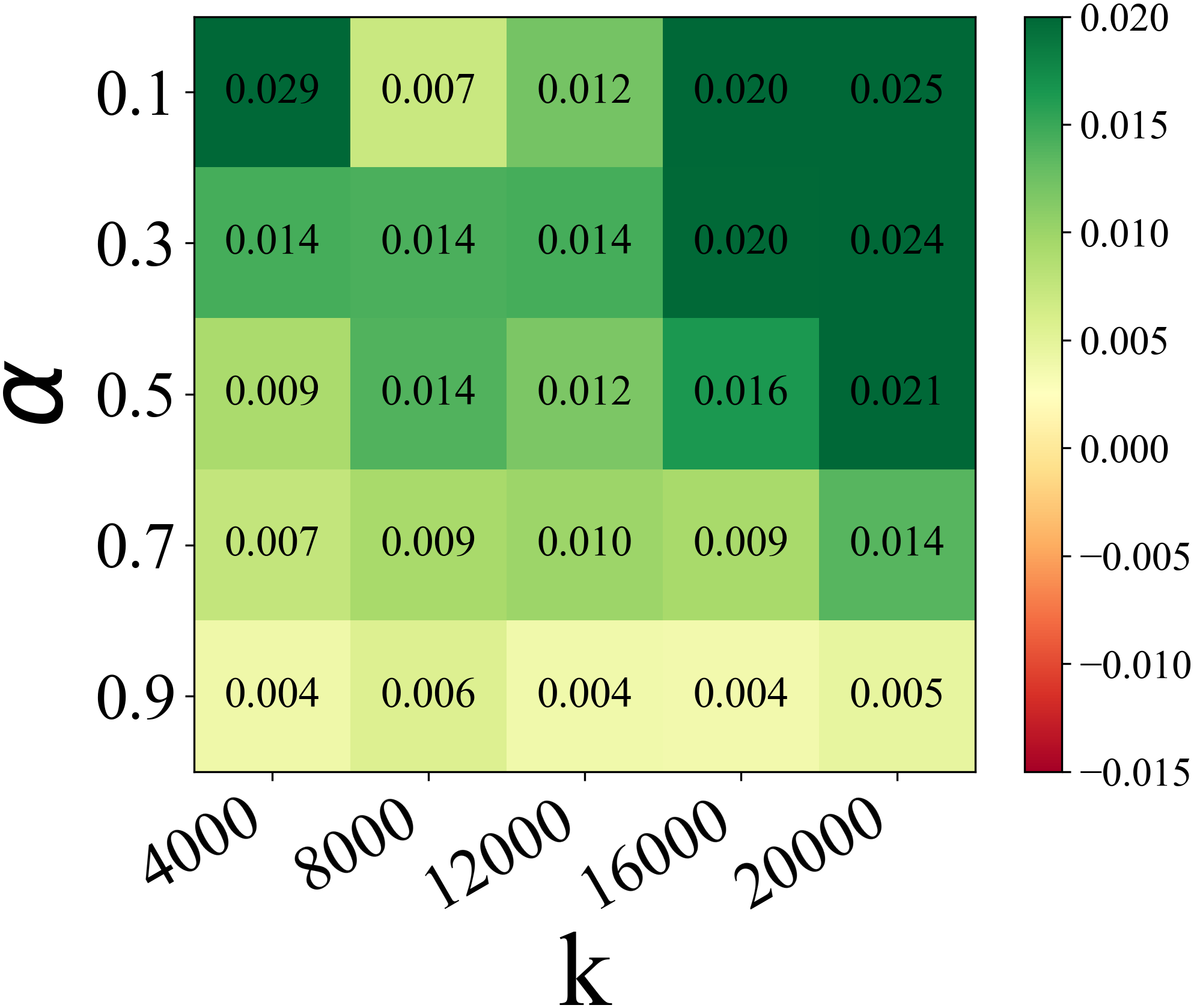}
        \vspace{-0.15in}
        \caption{DC (Borda)}
        \label{fig:grid_dc_borda}
    \end{subfigure}
    \hfill
    \begin{subfigure}[t]{0.245\linewidth}
        \centering
        \includegraphics[width=\linewidth]{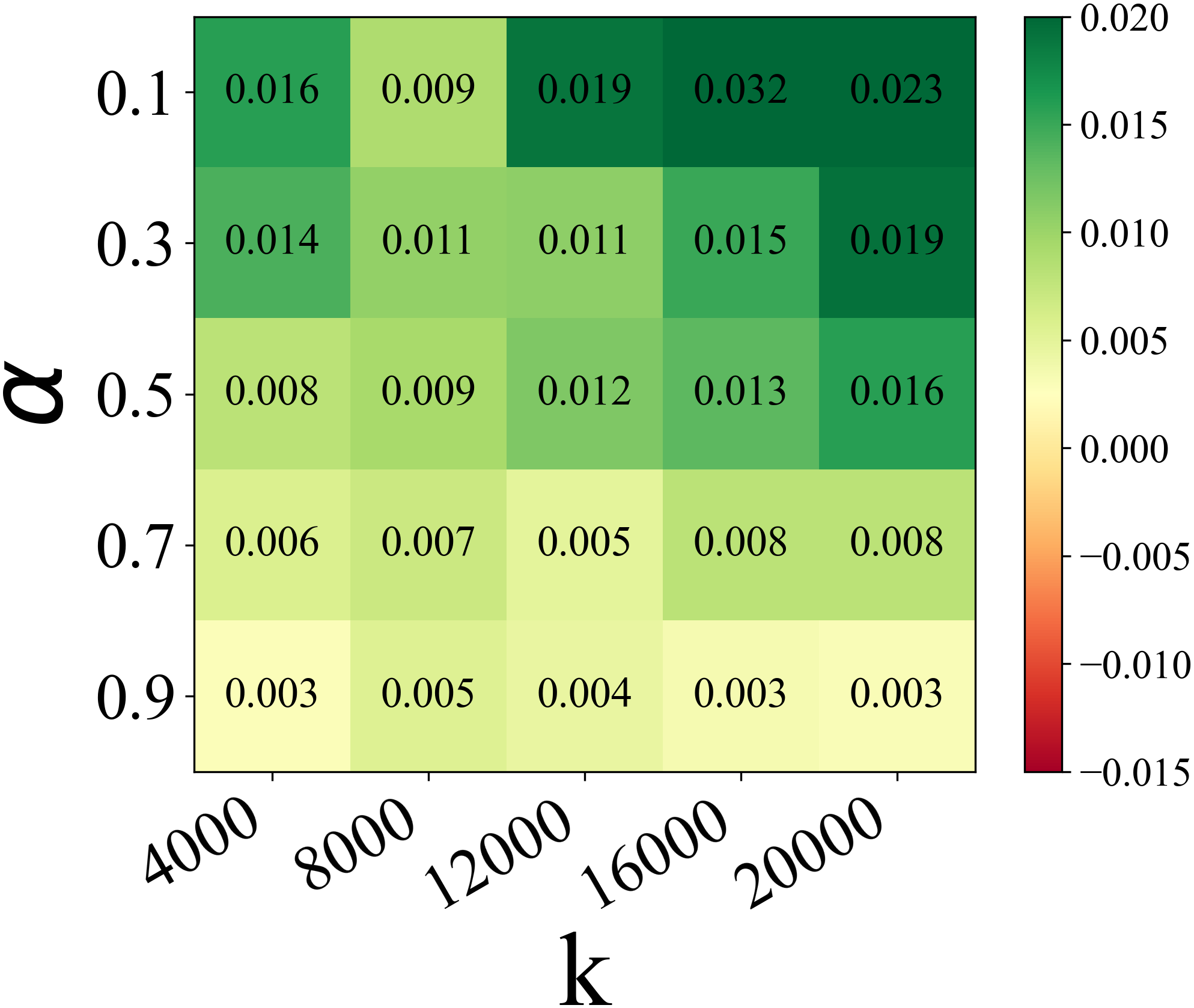}
        \vspace{-0.15in}
        \caption{SCert (Borda)}
        \label{fig:grid_sc_borda}
    \end{subfigure}
    \hfill
    \begin{subfigure}[t]{0.245\linewidth}
        \centering
        \includegraphics[width=\linewidth]{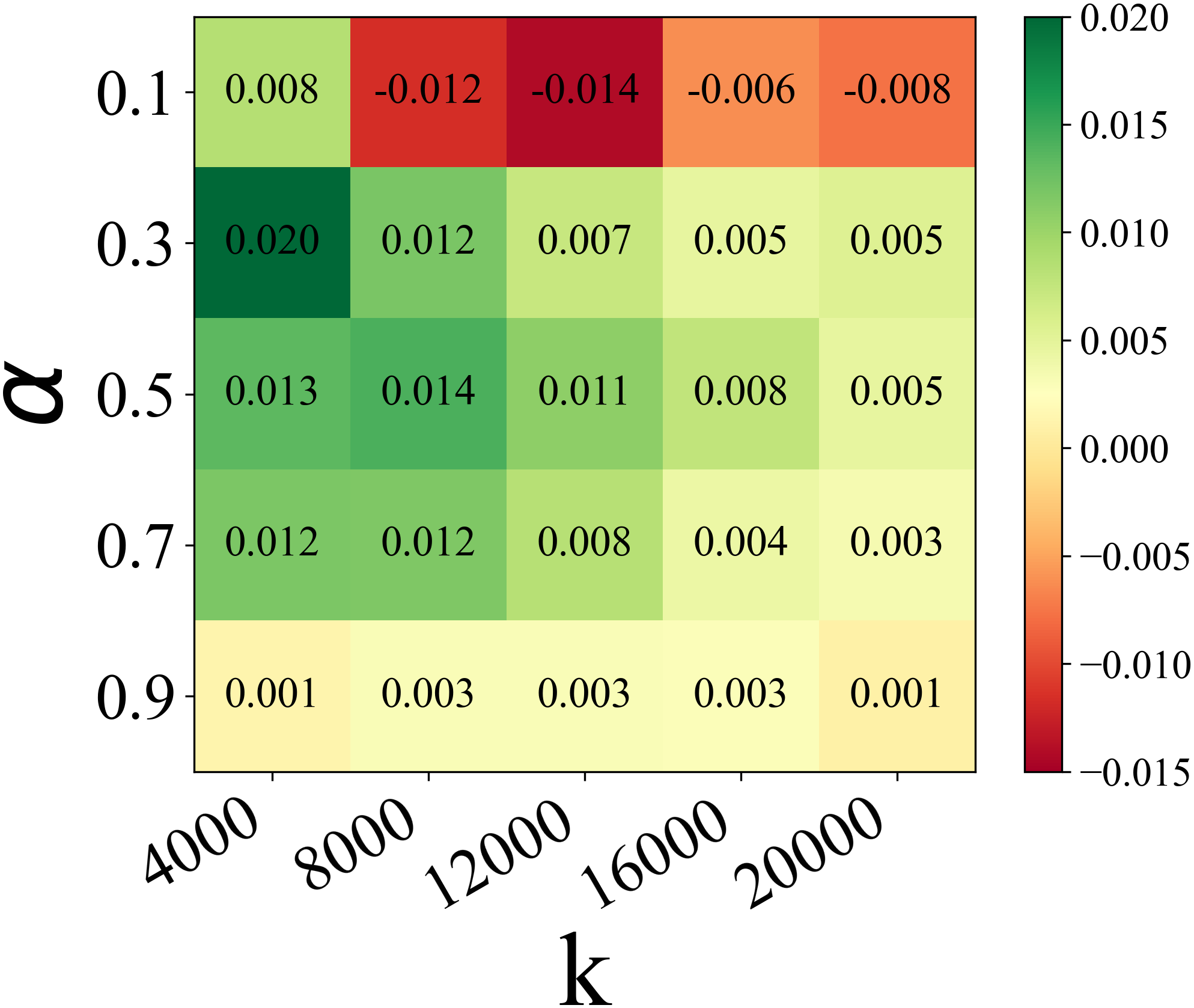}
        \vspace{-0.15in}
        \caption{DC (W-10\%)}
        \label{fig:grid_dc_w}
    \end{subfigure}
    \hfill
    \begin{subfigure}[t]{0.245\linewidth}
        \centering
        \includegraphics[width=\linewidth]{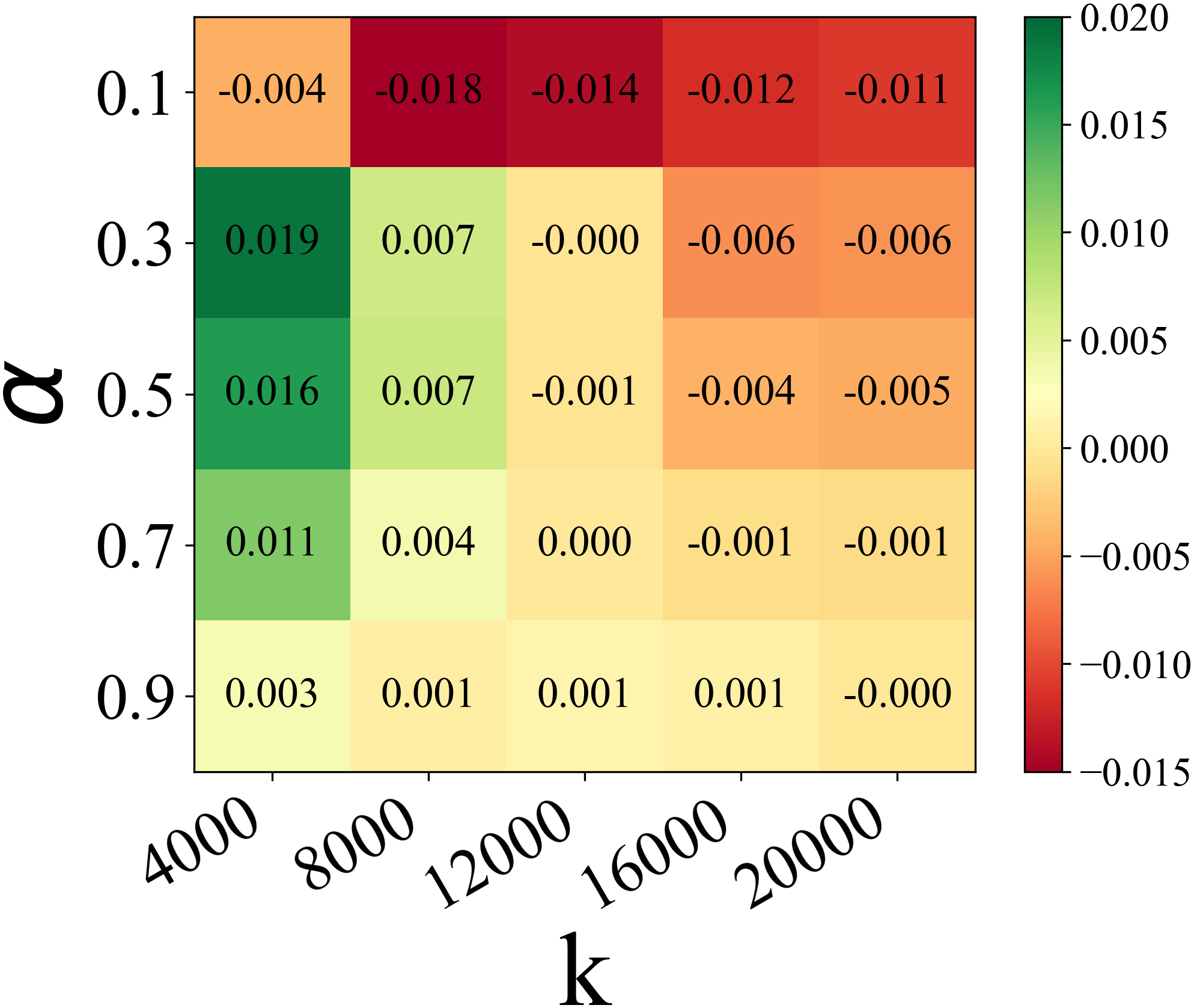}
        \vspace{-0.15in}
        \caption{SCert (W-10\%)}
        \label{fig:grid_sc_w}
    \end{subfigure}

    \caption{
    Effect of $k$ and $\alpha$ on Accuracy (Qwen, Math Datasets)
    }
    \vspace{-0.1in}
    \label{fig:grid}
\end{figure}

We analyze the effect of varying $\alpha$ and $k$. Consistent with our theoretical analysis, the optimal $\alpha$ should lie away from 0 and 1, and the optimal $k$ tends to occur at the boundary of the search range.
In Figure~\ref{fig:grid}, we evaluate $k \in \{4\text{k}, 8\text{k}, 12\text{k}, 16\text{k}, 20\text{k}\}$ and $\alpha \in \{0.1, 0.3, 0.5, 0.7, 0.9\}$ Each cell reports the change in accuracy relative to the vanilla, with positive values indicating improvement.

\textbf{Effect of $k$.} Across all four figures, the optimum occurs at either $k_{\text{min}}$ or $k_{\text{max}}$. However, $k_{\text{max}}$ is unstable—as shown in Figures~\ref{fig:grid_dc_w}--\ref{fig:grid_sc_w}—and the maximum possible $k$ varies unpredictably across samples, so we favor $k_{\text{min}}$. 
\textbf{Effect of $\alpha$.} For Borda (Figures~\ref{fig:grid_dc_borda}--\ref{fig:grid_sc_borda}), $\alpha = 0.3$ yields the largest gains. For W-10\% (Figures~\ref{fig:grid_dc_w}--\ref{fig:grid_sc_w}), $\alpha = 0.1$ performs best; we did not explore finer granularity below 0.1, because $\alpha \to 0$ is not desirable as Section~\ref{sec:hyper}. The key finding is that larger $\alpha \to 1$ is always worse, confirming that prefix-based path selection via {PreG} is consistently beneficial.

\subsection{Small budget $B$}
\label{sec:small_budget}




\begin{wrapfigure}{r}{0.27\textwidth}
    \centering
    \vspace{-25pt}
    \begin{minipage}{\linewidth}
        \centering
        \includegraphics[width=\linewidth]{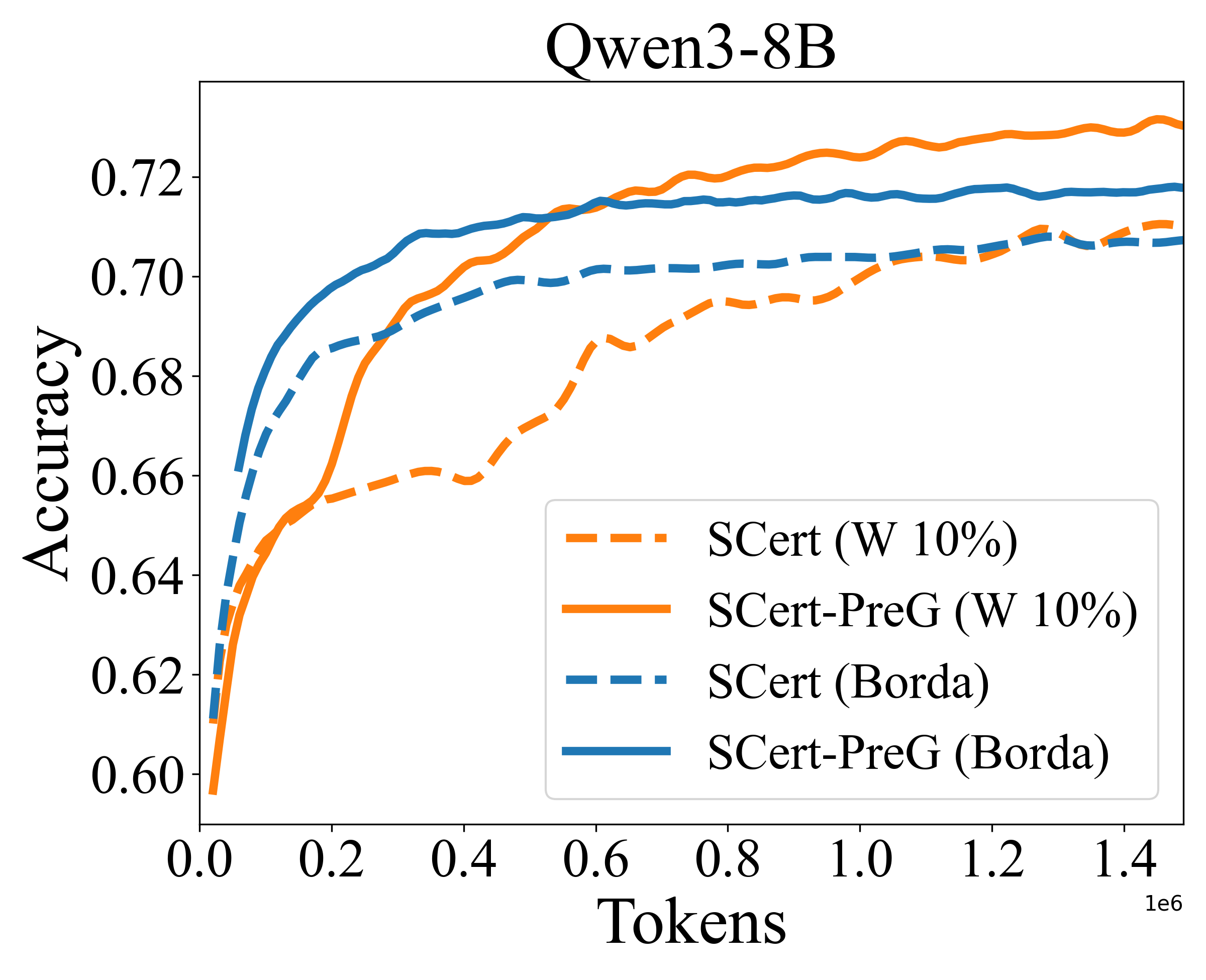}
        \vspace{-0.15in}
        \label{fig:small_qwen}
    \end{minipage}
    \begin{minipage}{\linewidth}
        \centering
        \includegraphics[width=\linewidth]{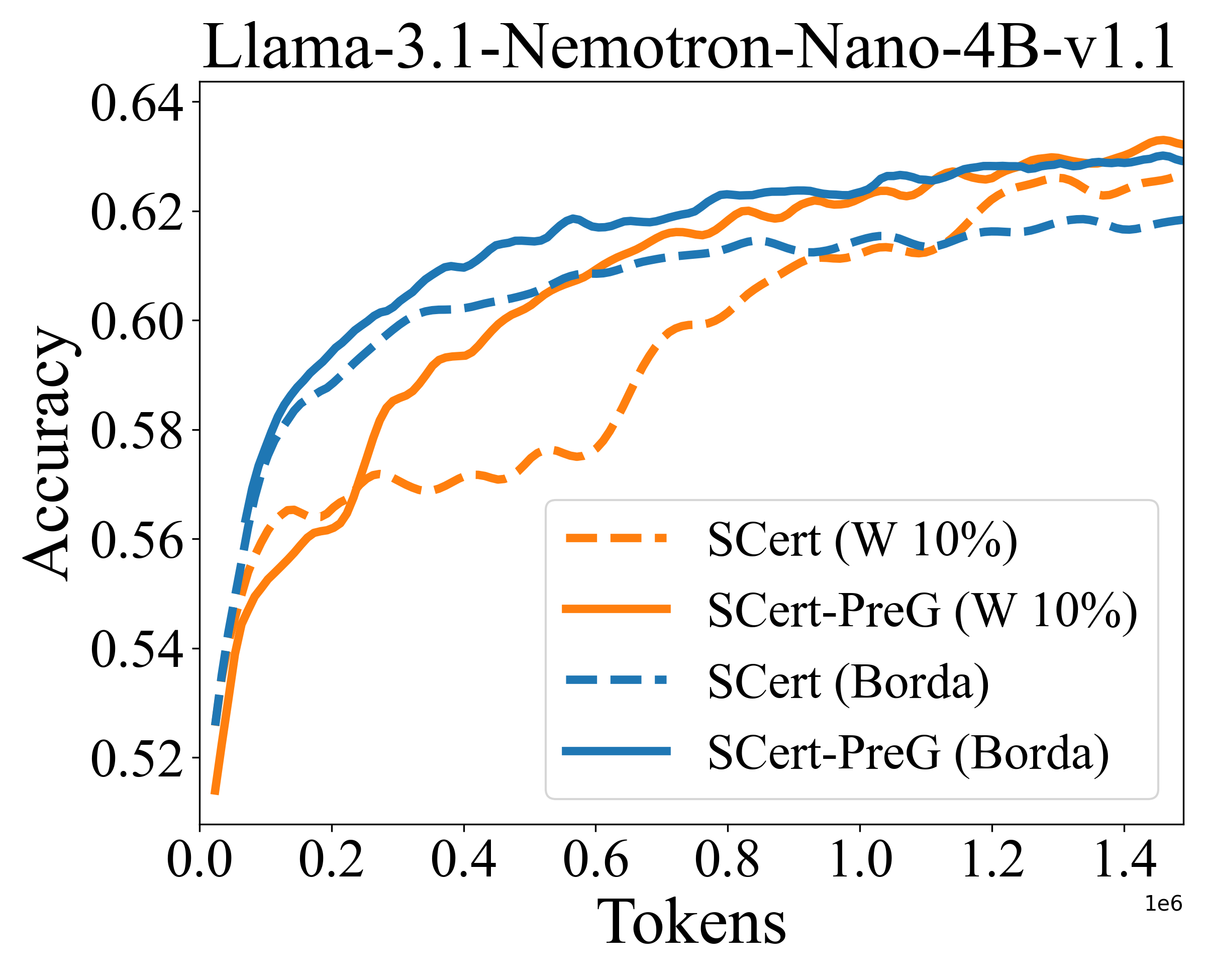}
        \vspace{-0.15in}
        \label{fig:small_llama}
    \end{minipage}
    \caption{Tokens vs. Acc.}
    \label{fig:small}
    \vspace{-20pt}
\end{wrapfigure}
To better illustrate how performance varies with budget, we plot the detailed accuracy curves in Figure~\ref{fig:small}, averaged across four math datasets.

For SCert, our method consistently outperforms the baseline across all budgets. For W-10\%, our method shows clear improvement beyond approximately 120k tokens for Qwen ($M_{\text{base}} \approx 6.48$) and 236k tokens for LLaMA ($M_{\text{base}} \approx 10.92$). These path counts are well below those used in typical parallel TTS settings. Also, as the curves show, the region below these budgets corresponds to the fastest phase of accuracy growth, where practitioners would rarely stop. Moreover, below these thresholds, PreG drops only marginally below the vanilla baseline.

An interesting observation is that the vanilla W-10\% exhibits a pronounced plateau between 200k--400k tokens where accuracy grows slowly, whereas our method largely eliminates this plateau. This directly supports our motivation for allocating budget to explore more prefixes: by surfacing high-quality candidates earlier, {PreG} accelerates convergence and removes the stagnation region.

\subsection{System-level Inference Efficiency}
\label{sec:multi_gpu}

We further evaluate the system-level efficiency of PreG
in a multi-GPU inference setting. Specifically, we use
Qwen3-8B with vLLM on four NVIDIA H200 GPUs, maintaining
the same distributed configuration for both methods.

\begin{table}[t]
    \centering
    \caption{Multi-GPU inference efficiency on four H200 GPUs
    using Qwen3-8B.}
    \label{tab:multi_gpu}
    \resizebox{0.65\columnwidth}{!}{%
        \begin{tabular}{lccccc}
            \toprule
            Method & Traces & Tokens & Latency
            & Avg. KV & Peak KV \\
            & per problem & (M) & (min)
            & (\%) & (\%) \\
            \midrule
            DC-Vanilla & 32 & 6.43 & 58.5 & 37.2 & 93.8 \\
            DC-PreG & $80 \rightarrow 24$ & 6.95
            & \textbf{40.6} & \textbf{31.3} & \textbf{74.7} \\
            \bottomrule
        \end{tabular}%
    }
\end{table}

As shown in Table~\ref{tab:multi_gpu}, DC-PreG reduces
end-to-end latency by 30.6\%, from 58.5 to 40.6 minutes.
Average KV-cache utilization decreases from 37.2\% to
31.3\%, while peak utilization decreases from 93.8\%
to 74.7\%. Notably, these improvements are achieved
despite generating 8.1\% more tokens overall.

This improvement is consistent with PreG's sequence-length
allocation: it explores more short prefixes but completes
fewer long trajectories, reducing long-lived KV-cache
occupancy and mitigating long-tail decoding overhead.
These results demonstrate that PreG's compute-allocation
benefits can translate into improved latency and
KV-cache efficiency in multi-GPU inference.


\section{Conclusions}
\label{sec:conclusion}


We proposed Prefix-Guided Sampling (PreG), a compute-efficient framework for parallel TTS that generates short prefixes, filters low-quality candidates via internal confidence scores, and completes only the most promising paths. Theoretically, we show that weak prefix signals can be amplified through aggregation near the decision boundary. Empirically, PreG consistently outperforms vanilla confidence-based TTS methods while reducing token usage by up to 80\%.

\textbf{Broader Impacts and Limitations.} This work may also enable more scalable misuse of LLMs, such as large-scale harmful content generation. Our method's effectiveness is bounded by the quality of the underlying confidence signal, limiting gains when the signal is weak. Under very small token budgets, PreG may not outperform the vanilla baseline.

\begin{ack}
This work was supported by Coefficient Giving. The authors have no competing interests to declare.
\end{ack}

\bibliographystyle{unsrtnat}
\bibliography{reference}

\newpage
\appendix

\section{Full proofs}
\label{appd:proof}

\subsection{Proof of Proposition~\ref{thm:margin-accuracy}}
\label{app:proof-margin-accuracy}

We prove Proposition~\ref{thm:margin-accuracy} in two steps. First, we show that the correctness of the final prediction is exactly characterized by the sign of the aggregated margin. Second, we derive a Gaussian approximation to the distribution of the margin.

\paragraph{Step 1: Exact characterization by the margin.}

The final prediction is defined as
\[
\hat y
=
\arg\max_y \mathrm{Conf}(y).
\]
By definition, the prediction is correct if and only if the score of the correct answer $y^*$ exceeds the score of every incorrect answer:
\[
\hat y=y^*
\quad
\Longleftrightarrow
\quad
\mathrm{Conf}(y^*) >
\max_{y\neq y^*}\mathrm{Conf}(y).
\]
Equivalently,
\[
\hat y=y^*
\quad
\Longleftrightarrow
\quad
\mathrm{Conf}(y^*)-\max_{y\neq y^*}\mathrm{Conf}(y)>0.
\]
Using the definition of the aggregated margin,
\[
\Delta
=
\mathrm{Conf}(y^*)-\max_{y\neq y^*}\mathrm{Conf}(y),
\]
we obtain
\[
\hat y=y^*
\quad
\Longleftrightarrow
\quad
\Delta>0.
\]
Therefore,
\[
\Pr(\hat y=y^*)
=
\Pr(\Delta>0).
\]

\paragraph{Step 2: Margin as a sum of signed per-path contributions.}

Let
\[
y_w
=
\arg\max_{y\neq y^*}\mathrm{Conf}(y)
\]
denote the strongest incorrect answer under the aggregation rule. Consider the signed contribution of the $i$-th path:
\[
Z_i
=
W(S_i)
\left(
\mathbf{1}[y_i=y^*]
-
\mathbf{1}[y_i=y_w]
\right).
\]
Then the difference between the aggregated score of the correct answer and that of the strongest incorrect answer can be written as
\[
\mathrm{Conf}(y^*)-\mathrm{Conf}(y_w)
=
\sum_{i=1}^{M}
W(S_i)
\left(
\mathbf{1}[y_i=y^*]
-
\mathbf{1}[y_i=y_w]
\right)
=
\sum_{i=1}^{M} Z_i.
\]
Since
\[
\Delta
=
\mathrm{Conf}(y^*)-\max_{y\neq y^*}\mathrm{Conf}(y)
=
\mathrm{Conf}(y^*)-\mathrm{Conf}(y_w),
\]
we have
\[
\Delta=\sum_{i=1}^{M} Z_i.
\]

\paragraph{Step 3: Mean and variance of the margin.}

By assumption,
\[
\mathbb{E}[Z_i]=\mu,
\qquad
\mathrm{Var}(Z_i)=\sigma^2.
\]
Since the $Z_i$ are independent, the mean and variance of $\Delta$ are
\[
\mathbb{E}[\Delta]
=
\mathbb{E}\left[\sum_{i=1}^{M}Z_i\right]
=
\sum_{i=1}^{M}\mathbb{E}[Z_i]
=
M\mu,
\]
and
\[
\mathrm{Var}(\Delta)
=
\mathrm{Var}\left(\sum_{i=1}^{M}Z_i\right)
=
\sum_{i=1}^{M}\mathrm{Var}(Z_i)
=
M\sigma^2.
\]

\paragraph{Step 4: Gaussian approximation.}

Because $Z_i$ are independent and bounded, the central limit theorem implies that for sufficiently large $M$,
\[
\frac{\Delta-M\mu}{\sqrt{M}\sigma}
\approx
\mathcal{N}(0,1).
\]
Equivalently,
\[
\Delta
\approx
\mathcal{N}(M\mu, M\sigma^2).
\]

\paragraph{Step 5: Approximate prediction accuracy.}

Using the exact characterization from Step 1,
\[
\Pr(\hat y=y^*)
=
\Pr(\Delta>0).
\]
Under the Gaussian approximation,
\[
\Pr(\Delta>0)
=
\Pr\left(
\frac{\Delta-M\mu}{\sqrt{M}\sigma}
>
\frac{-M\mu}{\sqrt{M}\sigma}
\right).
\]
Let
\[
Z
=
\frac{\Delta-M\mu}{\sqrt{M}\sigma}
\approx
\mathcal{N}(0,1).
\]
Then
\[
\Pr(\Delta>0)
\approx
\Pr\left(
Z>
-\frac{\sqrt{M}\mu}{\sigma}
\right).
\]
By symmetry of the standard normal distribution,
\[
\Pr\left(
Z>
-\frac{\sqrt{M}\mu}{\sigma}
\right)
=
\Phi\left(
\frac{\sqrt{M}\mu}{\sigma}
\right).
\]
Therefore,
\[
\Pr(\hat y=y^*)
\approx
\Phi\left(
\frac{\sqrt{M}\mu}{\sigma}
\right).
\]
This completes the proof.

\subsection{Derivation for Prefix-Length Selection}
\label{app:proof-boundary-k}

We prove that, under a linear margin growth assumption, the optimal prefix length is attained at one of the endpoints of the feasible interval.

\paragraph{Step 1: Objective.}

Recall that the effective voting strength is defined as
\[
V(k)
=
\sqrt{M(k)}\Delta(k).
\]

Under a fixed budget, the number of completed paths after prefix filtering is
\[
M(k)
=
\frac{\alpha B}{k+\alpha(T-k)}
=
\frac{\alpha B}{\alpha T+(1-\alpha)k}.
\]

Assume that the margin grows linearly with the prefix length:
\[
\Delta(k)=\Delta_0+\beta k.
\]

Therefore,
\[
V(k)
=
\sqrt{
\frac{\alpha B}{\alpha T+(1-\alpha)k}
}
(\Delta_0+\beta k).
\]

Since \(\sqrt{\alpha B}\) is independent of \(k\), maximizing \(V(k)\) is equivalent to maximizing
\[
\tilde V(k)
=
\frac{\Delta_0+\beta k}
{\sqrt{\alpha T+(1-\alpha)k}}.
\]

\paragraph{Step 2: Derivative.}

Let
\[
A(k)=\Delta_0+\beta k,
\qquad
C(k)=\alpha T+(1-\alpha)k.
\]

Then
\[
\tilde V(k)=A(k)C(k)^{-1/2}.
\]

Taking derivative with respect to \(k\),
\[
\tilde V'(k)
=
\beta C(k)^{-1/2}
-
\frac{1}{2}A(k)C(k)^{-3/2}(1-\alpha).
\]

Factoring out \(C(k)^{-3/2}>0\), we obtain
\[
\tilde V'(k)
=
C(k)^{-3/2}
\left[
\beta C(k)
-
\frac{1-\alpha}{2}A(k)
\right].
\]

Therefore, the sign of \(\tilde V'(k)\) is determined by
\[
G(k)
=
\beta C(k)
-
\frac{1-\alpha}{2}A(k).
\]

Substituting \(A(k)\) and \(C(k)\), we get
\[
G(k)
=
\beta\left(\alpha T+(1-\alpha)k\right)
-
\frac{1-\alpha}{2}\left(\Delta_0+\beta k\right).
\]

Equivalently,
\[
G(k)
=
\alpha\beta T
+
\frac{1-\alpha}{2}\beta k
-
\frac{1-\alpha}{2}\Delta_0.
\]

Thus \(G(k)\) is an affine function of \(k\).

\paragraph{Step 3: Shape of the objective.}

Since \(G(k)\) is affine in \(k\), the derivative \(\tilde V'(k)\) can change sign at most once over the interval
\[
k\in[k_{\min},k_{\max}].
\]

Therefore, \(\tilde V(k)\) is either monotone over the interval, or it decreases first and then increases. In either case, \(\tilde V(k)\) cannot have a strict interior maximum.

Hence the maximum of \(\tilde V(k)\), and therefore the maximum of \(V(k)\), must be attained at one of the two endpoints:
\[
k^*\in\{k_{\min},k_{\max}\}.
\]


\section{Definitions of Used Confidence Score in this Work}
\label{appd:score_def}


Let $S = (s_1, s_2, \ldots, s_T)$ be a generated reasoning path, where each $s_t \in \mathcal{V}$ is a token. Let $p_t(v) = P(s_t = v \mid s_{<t})$ denote the model's predicted probability distribution over vocabulary $\mathcal{V}$ at position $t$, and let $|\mathcal{V}|$ denote the vocabulary size.

\paragraph{Likelihood (LL).}
The likelihood score aggregates the log-probabilities of the generated tokens:
\begin{equation}
    \mathrm{LL}(S) = \mathrm{exp}(\frac{1}{T} \sum_{t=1}^{T} \log p_t(s_t)).
\end{equation}

Exp makes sure LL is not negative. If it negative it cannot be used for weighted voting.

\paragraph{Entropy.}
The entropy score measures the average uncertainty of the model's predictions, with lower entropy indicating higher confidence:
\begin{equation}
    \mathrm{Entropy}(S) = -\frac{1}{T} \sum_{t=1}^{T} \sum_{v \in \mathcal{V}} p_t(v) \log p_t(v).
\end{equation}
A lower entropy corresponds to a higher-confidence path; in practice, the negative entropy $\mathrm{exp}(-\mathrm{Entropy}(S))$ is used as the confidence score.

\paragraph{Gini Impurity.}
Gini Impurity measures the concentration of the predicted distribution, analogous to the Gini coefficient:
\begin{equation}
    \mathrm{Gini}(S) = \frac{1}{T} \sum_{t=1}^{T} \left(1 - \sum_{v \in \mathcal{V}} p_t(v)^2\right).
\end{equation}
A lower Gini Impurity indicates a more peaked (confident) distribution; the negative value $-\mathrm{Gini}(S)$ is used as the confidence score.

\paragraph{Diff-token.}
Diff-token measures the gap between the top-1 and top-2 predicted probabilities at each step, rewarding distributions where one token is clearly preferred:
\begin{equation}
    \mathrm{DiffToken}(S) = \frac{1}{T} \sum_{t=1}^{T} \left( p_t^{(1)} - p_t^{(2)} \right),
\end{equation}
where $p_t^{(1)} \geq p_t^{(2)}$ are the largest and second-largest probabilities at position $t$.

\paragraph{Conf-Lens.}
Conf-Lens is a new confidence score we introduce, derived from the logit 
lens~\cite{logitlens2020}. The logit lens projects intermediate hidden 
states $h_t^{(\ell)}$ at layer $\ell$ through the final unembedding matrix 
to obtain an intermediate predicted probability $\tilde{p}_t^{(\ell)}(s_t)$ 
for the generated token $s_t$. We measure confidence by aggregating these 
intermediate probabilities over a set of later layers $\mathcal{L}$:
\begin{equation}
    \mathrm{ConfLens}(S) = \exp\!\left(\frac{1}{T} \sum_{t=1}^{T} 
    \log\left(\frac{1}{|\mathcal{L}|} \sum_{\ell \in \mathcal{L}} 
    \tilde{p}_t^{(\ell)}(s_t)\right)\right),
\end{equation}
where $\mathcal{L}$ denotes the set of later layers used. This can be 
interpreted as the geometric mean over token positions of the 
layer-averaged intermediate probability assigned to each generated token. 
A higher score indicates that the later intermediate layers consistently 
assign high probability to the generated tokens, reflecting greater 
internal agreement and confidence. We include Conf-Lens only to evaluate whether PreG generalizes across different types of confidence signals, rather than as a core contribution.

\paragraph{DeepConf (DC).}
DeepConf measures the concentration of predicted distributions by aggregating token-level confidence based on the peakedness of the softmax output:
\begin{equation}
    \mathrm{DC}(S) = \frac{1}{T} \sum_{t=1}^{T} \left( -\frac{1}{k}\sum_{j=1}^k\log p_t(j) \right).
\end{equation}

\paragraph{Self-Certainty (SCert).}
Self-Certainty is a KL-divergence-inspired distributional confidence metric defined as:
\begin{equation}
    \mathrm{SCert}(S) = -\frac{1}{TV} \sum_{t=1}^{T} \sum_{j=1}^{V} \log \bigl(V \cdot p_t(j)\bigr),
\end{equation}
where $V = |\mathcal{V}|$ is the vocabulary size and $p_t(j)$ denotes the predicted probability 
of the $j$-th vocabulary token at position $t$. This is equivalent to the KL divergence 
$D_{\mathrm{KL}}(p_t \| \mathcal{U})$ up to a constant, where $\mathcal{U}$ is the uniform 
distribution over $\mathcal{V}$.



\section{Licenses of Assets}
\label{appd:license}

\begin{table}[t]
    \centering
   \caption{License information of assets}
    \resizebox{1\textwidth}{!}{ 
    \begin{tabular}{lll}
        \toprule
        Asset & License & Link \\
        \midrule
        Qwen/Qwen3-8B & Apache license 2.0 & https://huggingface.co/Qwen/Qwen3-8B \\
        nvidia/Llama-3.1-Nemotron-Nano-4B-v1.1 & nvidia-open-model-license & https://huggingface.co/nvidia/Llama-3.1-Nemotron-Nano-4B-v1.1 \\
        Qwen/Qwen2.5-Math-PRM-7B & qwen & https://huggingface.co/Qwen/Qwen2.5-Math-PRM-7B \\
        Skywork/Skywork-Reward-V2-Qwen3-8B & Apache license 2.0 & https://huggingface.co/Skywork/Skywork-Reward-V2-Qwen3-8B \\
        
        lchen001/AIME2025 & Apache license 2.0 & https://huggingface.co/datasets/lchen001/AIME2025 \\

        MathArena/brumo\_2025 & CC-BY-NC-SA-4.0 & https://huggingface.co/datasets/MathArena/brumo\_2025 \\

        MathArena/hmmt\_feb\_2025 & CC-BY-NC-SA-4.0 & https://huggingface.co/datasets/MathArena/hmmt\_feb\_2025 \\

        MathArena/cmimc\_2025 & CC-BY-NC-SA-4.0 & https://huggingface.co/datasets/MathArena/cmimc\_2025 \\

        Idavidrein/gpqa & CC-BY-4.0 & https://huggingface.co/datasets/Idavidrein/gpqa \\
        
        \bottomrule
    \end{tabular}
    }
    \label{tab:license}
\end{table}

In Table~\ref{tab:license}, we present the license information of all the assets including the data resources and the model we use.

\section{Additional Experiment Results}
\label{appd:add_exp}

\subsection{Studies on the internal signals from prefix}
\label{appd:prefix_score}

\begin{figure}[t]
\centering

\begin{minipage}[t]{0.66\linewidth}
    \centering

    \begin{subfigure}[t]{0.49\linewidth}
        \centering
        \includegraphics[width=\linewidth]{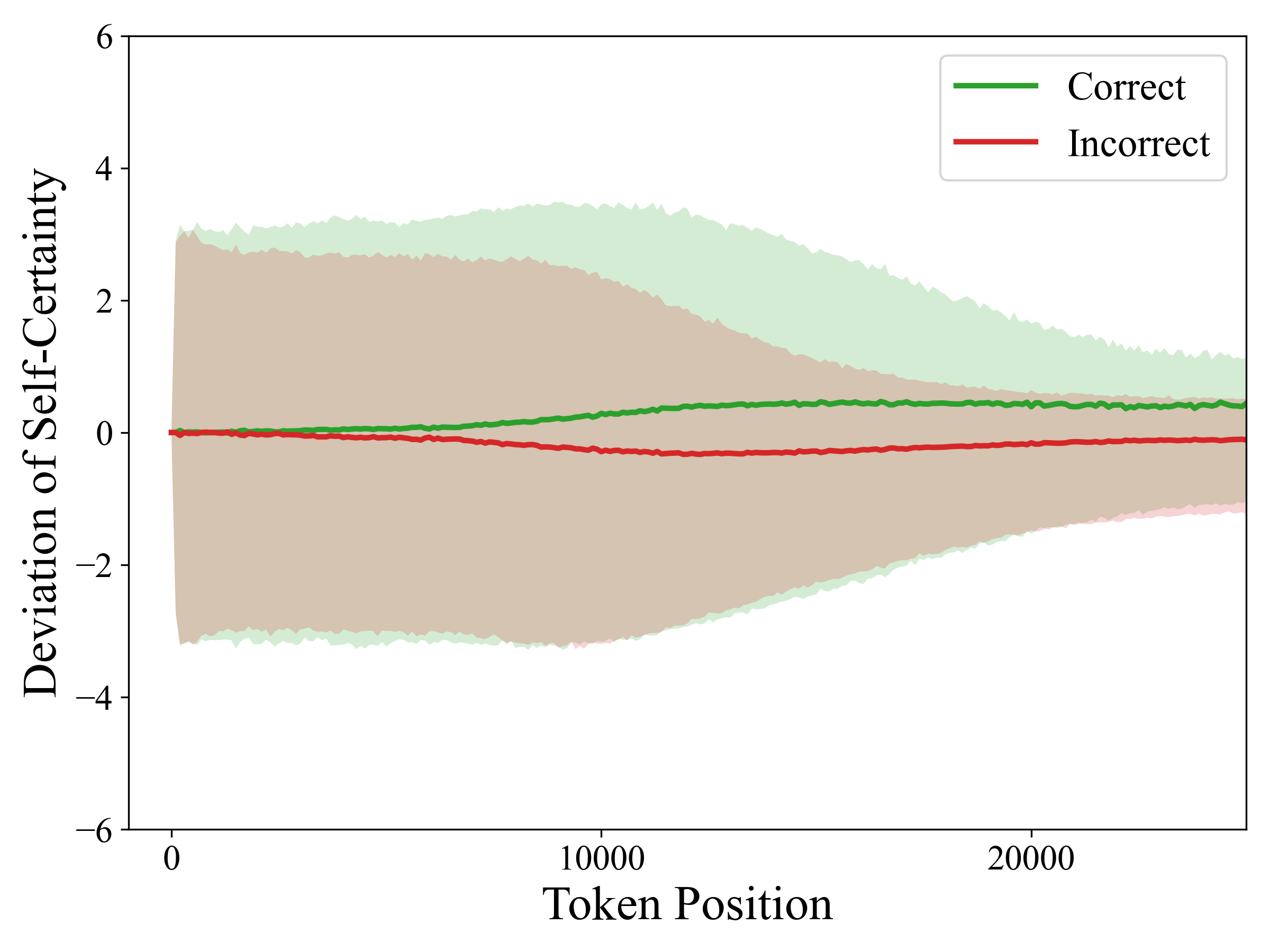}
        \vspace{-0.23in}
        \caption{}
        \label{fig:score_curve_begin}
    \end{subfigure}
    \hfill
    \begin{subfigure}[t]{0.49\linewidth}
        \centering
        \includegraphics[width=\linewidth]{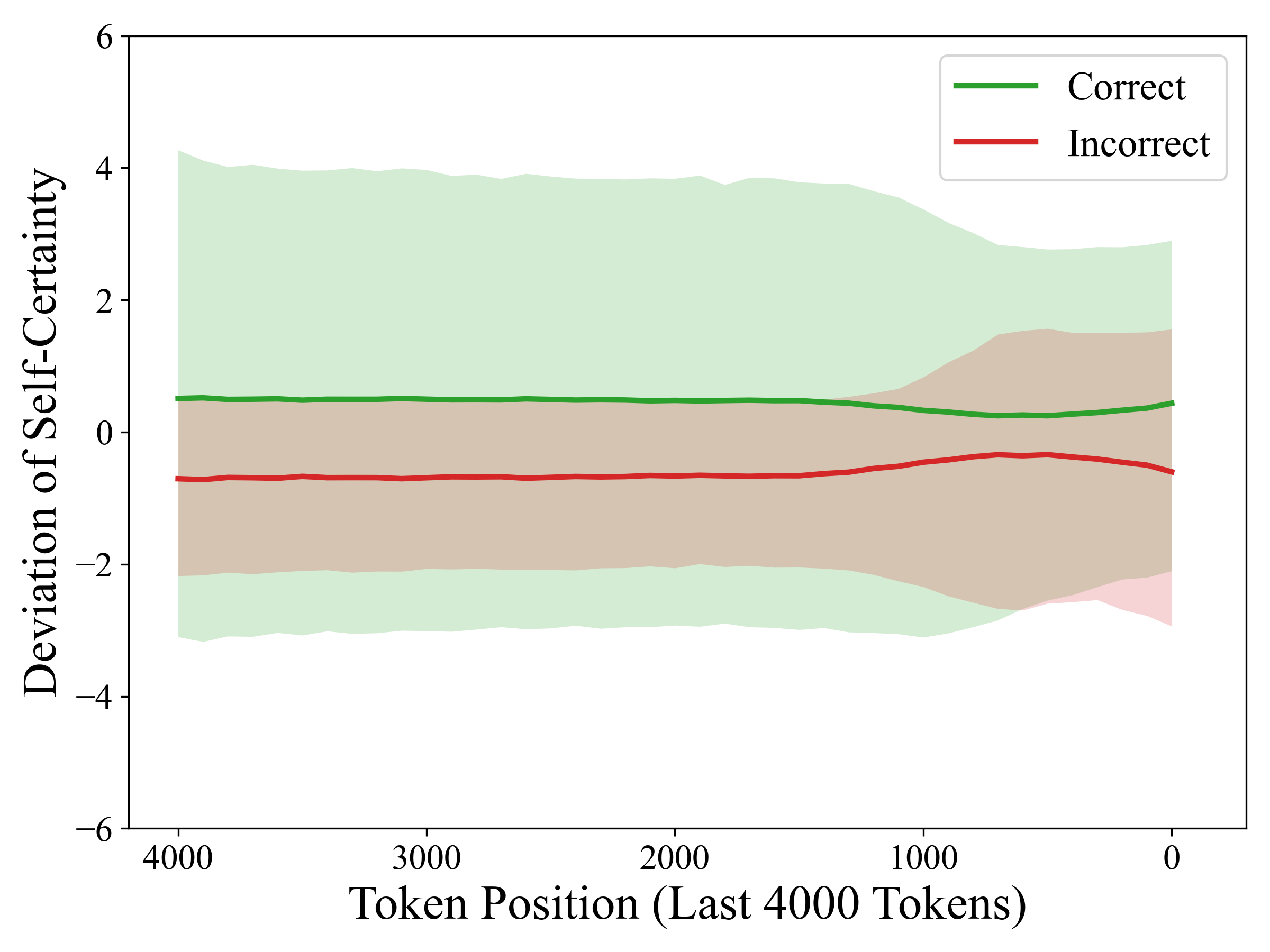}
        \vspace{-0.23in}
        \caption{}
        \label{fig:score_curve_end}
    \end{subfigure}

    \vspace{-0.07in}
    \caption{
    Results on prefix scores.
    }
    \label{fig:score_curve}
\end{minipage}
\hfill
\begin{minipage}[t]{0.325\linewidth}
    \centering

    \includegraphics[width=\linewidth]{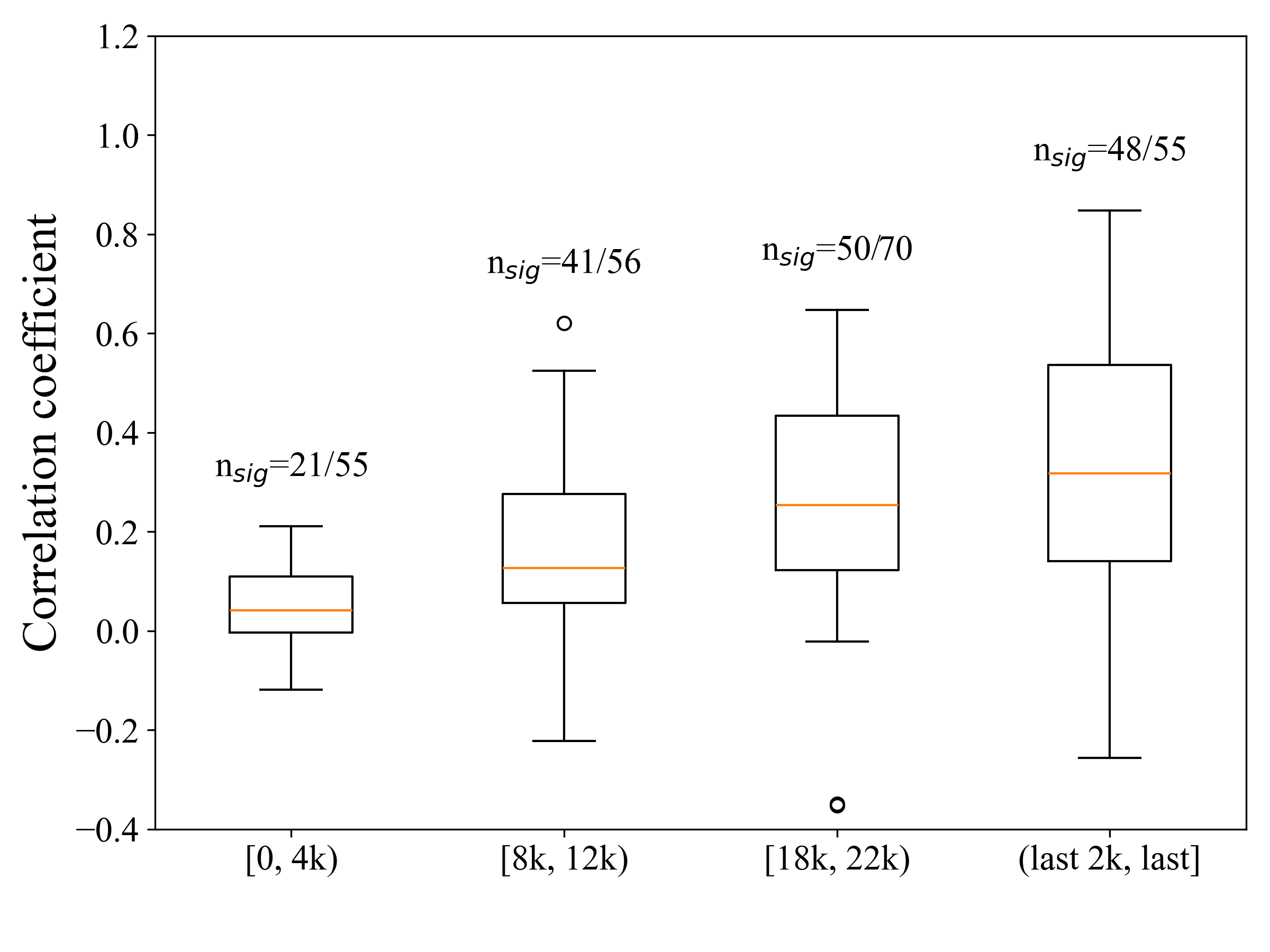}

    \vspace{0.07in}
    \caption{
    Correlation coefficients
    }
    \label{fig:r}
\end{minipage}

\end{figure}

Similar to Self-Consistency, parallel methods which uses token-level confidence scores typically make decisions, such as evaluating path quality or aggregating paths, only after full path generation. Many early truncation methods also rely on signals from completed paths rather than prefix-based stopping within a single path.
While full-path signals are more informative, they incur unnecessary cost on paths that yield incorrect answers. 
In this subsection, we conduct a preliminary study to explore how much prefix information can help reduce unnecessary tokens. 

We first examine how confidence evolves along correct and incorrect paths. In Figure~\ref{fig:score_curve}, we plot the Self-Certainty of Qwen3 on four different math datasets. For each question, we sample 512 responses from Qwen3-8B and compute token-level Self-Certainty, enabling a comparison between paths that lead to correct versus incorrect final answers. Since such comparisons are most meaningful within the same question, we normalize each token’s score by subtracting the mean score at that token position across all samples for that question. Figure~\ref{fig:score_curve_begin} shows the variation from the first token onward, while Figure~\ref{fig:score_curve_end} presents the results over the last 4,000 tokens.

The average length of Qwen3 reasoning paths on these problems is 2.1k tokens. From Figure~\ref{fig:score_curve_begin}, we observe that Self-Certainty shows almost no difference at the very beginning. Starting from around 2,000 tokens, a noticeable gap in the mean emerges, which continues to widen until approximately 12,000 tokens. From Figure~\ref{fig:score_curve_end}, we further observe that the mean difference becomes larger in the final 4,000 tokens. This suggests that, although the difference is small, early tokens have already begun to diverge and start to provide informative signals that can distinguish potential correct and incorrect paths.

Compared to Figure~\ref{fig:score_curve}, Figure~\ref{fig:r} further illustrates how Self-Certainty can help identify correct answers. Following the use of internal signals, we aggregate token-level information by averaging over different segments, i.e., token ranges [0, 4k), [8k, 12k), [18k, 22k), and [last2k, last]. We then compute, for each question, the correlation between sample-level scores and labels, yielding the relationship between samples and confidence scores.

In Figure~\ref{fig:r}, after removing questions with highly imbalanced labels, we present box plots of the correlation coefficients across all questions, along with the proportion of significant correlations (e.g., $n_{\text{sig}}$). We observe that later segments exhibit stronger correlations with the labels; however, even within the first 4k tokens, a clear correlation already emerges. Approximately 75\% of questions show a positive correlation between confidence scores and labels in this early segment, with 38\% ($n_{\text{sig}} = 21/55$) being statistically significant. These findings suggest that, even from prefixes, it is possible to select path candidates more likely to yield correct final answers, thereby improving overall performance.

\subsection{STD of Main Results}
\label{appd:std_results}

\begin{table}
  \caption{STD}
  \label{tab:std}
  \centering
  \resizebox{\linewidth}{!}{
  \begin{tabular}{clccccc}
    \toprule
    \multicolumn{2}{c}{\multirow{2}{*}{Qwen3-8B}} & AIME25 & BRUMO & HMMT & CMIMC & GPQA-D \\
     & & (100k/300k/1000k) & (100k/300k/1000k) & (100k/300k/1000k) & (100k/300k/1000k) & (100k/300k/1000k) \\

    \midrule

    \multicolumn{2}{c}{Pass@1} & 67.94 & 69.64 & 45.30 & 46.14 & 49.80 \\


    \multicolumn{2}{c}{Majority voting} & 3.30 / 2.69 / 2.68 & 3.67 / 2.98 / 2.33 & 4.88 / 3.06 / 2.31 & 4.03 / 3.02 / 1.52 & 1.77 / 1.25 / NA \\

    \multicolumn{2}{c}{PRM} & 2.64 / 2.81 / 2.89 & 3.48 / 2.78 / 2.09 & 3.62 / 2.96 / 1.96 & 3.36 / 3.11 / 1.83 & 1.83 / 1.35 / 1.09 \\
    \multicolumn{2}{c}{ESC} & 2.69 / 2.78 / 2.47 & 3.58 / 2.61 / 2.03 & 3.92 / 2.94 / 2.59 & 3.26 / 2.35 / 1.52 & 1.58 / 1.29 / NA \\
    \multicolumn{2}{c}{ST-BoN} & 3.82 / 2.98 / 2.33 & 3.51 / 3.13 / 2.43 & 4.58 / 3.60 / 2.77 & 4.43 / 3.16 / 2.04 & 1.75 / 1.27 / 1.07 \\
    \multicolumn{2}{c}{CGES} & 4.66 / 4.12 / NA & 4.15 / 3.24 / NA & 5.34 / 4.29 / NA & 4.58 / 3.85 / NA & NA / NA / NA \\

    \midrule

    \multirow{6}{*}{Borda} & LL & 3.27 / 2.33 / 2.62 & 4.06 / 2.46 / 1.74 & 4.62 / 2.87 / 2.59 & 3.99 / 3.80 / 2.15 & 1.88 / 1.56 / 1.02 \\
    & LL-PreG & 3.41 / 3.19 / 2.80 & 3.32 / 2.07 / 1.25 & 4.74 / 3.50 / 3.16 & 4.02 / 3.83 / 2.57 & 1.45 / 1.23 / 0.48 \\
    & DC & 3.78 / 2.89 / 2.72 & 3.43 / 2.65 / 2.13 & 4.69 / 3.21 / 1.89 & 4.08 / 3.85 / 3.17 & 1.59 / 1.36 / 0.90 \\
    & DC-PreG & 3.23 / 2.87 / 1.70 & 2.88 / 1.93 / 0.95 & 4.58 / 3.04 / 2.07 & 4.36 / 3.26 / 2.72 & 1.41 / 1.18 / 0.34 \\
    & SCert & 3.45 / 3.07 / 2.66 & 3.83 / 2.59 / 1.73 & 5.19 / 3.44 / 2.08 & 3.73 / 3.65 / 2.87 & 1.75 / 1.25 / 0.86 \\
    & SCert-PreG & 3.01 / 3.02 / 1.97 & 3.29 / 1.68 / 0.47 & 4.85 / 3.42 / 2.05 & 3.95 / 3.20 / 3.04 & 1.57 / 1.21 / 0.37 \\

    \midrule

    \multirow{6}{*}{W-10\%}   & LL & 4.93 / 5.20 / 3.54 & 4.65 / 4.65 / 3.32 & 4.69 / 4.83 / 3.78 & 4.91 / 4.72 / 3.76 & 2.56 / 2.20 / 1.41 \\
  & LL-PreG & 5.79 / 3.71 / 2.72 & 4.62 / 3.24 / 2.24 & 5.80 / 4.47 / 3.71 & 4.78 / 4.53 / 3.62 & 2.34 / 1.62 / 1.42 \\
  & DC & 4.66 / 4.58 / 2.72 & 4.41 / 4.50 / 2.95 & 4.98 / 5.35 / 3.57 & 4.71 / 4.46 / 3.50 & 2.34 / 1.93 / 1.30 \\
  & DC-PreG & 4.33 / 3.41 / 1.84 & 4.05 / 2.53 / 1.62 & 5.28 / 5.09 / 3.09 & 4.83 / 4.37 / 3.39 & 2.29 / 1.80 / 1.24 \\
  & SCert & 4.86 / 4.83 / 2.83 & 4.50 / 4.35 / 2.70 & 5.15 / 4.96 / 3.62 & 4.73 / 4.58 / 4.11 & 2.28 / 1.84 / 1.30 \\
  & SCert-PreG & 4.57 / 3.77 / 2.21 & 3.72 / 2.68 / 1.69 & 5.30 / 4.56 / 2.87 & 5.02 / 4.55 / 3.58 & 2.49 / 1.80 / 1.20 \\

    \midrule
    \multicolumn{2}{c}{\multirow{2}{*}{Llama-Nemotron-4B}} & AIME25 & BRUMO & HMMT & CMIMC & GPQA-D \\
     & & (100k/300k/1000k) & (100k/300k/1000k) & (100k/300k/1000k) & (100k/300k/1000k) & (100k/300k/1000k) \\

    \midrule

    \multicolumn{2}{c}{Pass@1} & 56.41 & 63.76 & 41.11 & 37.88 & 43.03 \\

    \multicolumn{2}{c}{Majority voting} & 4.62 / 3.46 / 2.51 & 3.56 / 3.36 / 2.53 & 4.07 / 3.28 / 2.77 & 4.04 / 2.83 / 2.09 & 2.37 / 1.54 / NA \\
    \multicolumn{2}{c}{PRM} & 2.64 / 2.81 / 2.89 & 3.48 / 2.78 / 2.09 & 3.62 / 2.96 / 1.96 & 3.36 / 3.11 / 1.83 & 1.83 / 1.35 / 1.09 \\
    \multicolumn{2}{c}{ESC} & 2.69 / 2.78 / 2.47 & 3.58 / 2.61 / 2.03 & 3.92 / 2.94 / 2.59 & 3.26 / 2.35 / 1.52 & 1.58 / 1.29 / NA \\
    \multicolumn{2}{c}{ST-BoN} & 3.82 / 2.98 / 2.33 & 3.51 / 3.13 / 2.43 & 4.58 / 3.60 / 2.77 & 4.43 / 3.16 / 2.04 & 1.75 / 1.27 / 1.07 \\
    \multicolumn{2}{c}{CGES}  & 4.66 / 4.12 / NA & 4.15 / 3.24 / NA & 5.34 / 4.29 / NA & 4.58 / 3.85 / NA & NA / NA / NA \\

    \midrule

    \multirow{6}{*}{Borda}   & LL & 4.08 / 3.43 / 2.63 & 3.58 / 3.69 / 3.03 & 4.67 / 3.56 / 3.04 & 3.62 / 2.87 / 1.76 & 2.56 / 1.77 / 1.21 \\
  & LL-PreG & 4.85 / 3.63 / 2.60 & 3.45 / 3.36 / 2.83 & 4.59 / 2.79 / 2.38 & 4.05 / 3.19 / 1.81 & 2.53 / 1.55 / 1.09 \\
  & DC & 3.75 / 3.69 / 2.57 & 3.74 / 3.68 / 2.87 & 4.80 / 3.46 / 3.09 & 3.42 / 2.87 / 1.70 & 2.51 / 1.88 / 1.25 \\
  & DC-PreG & 4.36 / 3.87 / 2.40 & 3.30 / 3.29 / 3.05 & 4.25 / 3.87 / 2.35 & 4.48 / 2.93 / 1.51 & 2.44 / 1.43 / 1.08 \\
  & SCert & 3.86 / 3.64 / 2.56 & 3.30 / 3.52 / 3.02 & 4.76 / 3.54 / 3.32 & 3.38 / 2.94 / 1.90 & 2.42 / 1.75 / 1.31 \\
  & SCert-PreG & 4.17 / 3.84 / 2.03 & 3.34 / 3.22 / 2.80 & 4.38 / 3.23 / 2.51 & 4.01 / 3.25 / 1.74 & 2.17 / 1.44 / 0.99 \\

    \midrule
    
    \multirow{6}{*}{W-10\%}   & LL & 5.33 / 5.29 / 4.22 & 4.23 / 4.40 / 3.06 & 5.11 / 4.89 / 3.66 & 4.15 / 4.34 / 3.94 & 2.63 / 2.76 / 2.19 \\
  & LL-PreG & 5.66 / 4.04 / 3.20 & 4.05 / 3.54 / 3.18 & 4.82 / 4.01 / 2.63 & 4.91 / 3.27 / 3.08 & 2.43 / 2.09 / 1.53 \\
  & DC & 5.11 / 4.98 / 4.02 & 3.52 / 3.66 / 3.01 & 4.87 / 4.92 / 3.48 & 3.94 / 4.49 / 4.01 & 2.72 / 2.72 / 2.09 \\
  & DC-PreG & 5.77 / 4.92 / 3.21 & 3.70 / 3.75 / 3.13 & 4.61 / 4.52 / 2.94 & 5.09 / 4.09 / 2.97 & 2.38 / 2.38 / 1.80 \\
  & SCert & 4.77 / 4.83 / 4.30 & 3.73 / 3.92 / 2.96 & 4.86 / 4.98 / 3.75 & 3.97 / 4.98 / 3.92 & 2.65 / 2.39 / 1.99 \\
  & SCert-PreG & 5.46 / 4.43 / 2.96 & 4.04 / 3.72 / 3.51 & 5.01 / 4.28 / 2.91 & 4.64 / 4.13 / 2.81 & 2.29 / 2.02 / 1.71 \\

    \bottomrule
  \end{tabular}
  }
\end{table}

The STD of main results are reported in Table~\ref{tab:std}.

\subsection{Comparison with Other Aggregation Methods and Confidence Scores}
\label{appd:other_vote_score}

\begin{table}
  \caption{Other aggregation methods}
  \label{tab:diff_vote}
  \centering
  \begin{tabular}{clc}
    \toprule
    Qwen & & Math \\

    \midrule

\multirow{6}{*}{Weighted voting} & LL & 64.61 / 67.29 / 68.27 \\
& LL-PreG & 62.63 / 67.75 / 69.59 \\
& DC & 65.71 / 67.90 / 68.53 \\
& DC-PreG & 64.84 / 68.59 / 69.96 \\
& SCert & 65.97 / 68.01 / 68.76 \\
& SCert-PreG & 65.42 / 68.74 / 70.38 \\

\midrule

& Majority voting & 64.65 / 67.18 / 68.24 \\
\multirow{3}{*}{Majority voting} & LL-PreG & 62.51 / 67.66 / 69.54 \\
& DC-PreG & 62.72 / 68.08 / 69.71 \\
& SCert-PreG & 62.83 / 67.93 / 69.86 \\

    \bottomrule
  \end{tabular}
\end{table}

\begin{table}
  \caption{Other confidences scores}
  \label{tab:other_score}
  \centering
  \begin{tabular}{clcccccc}
    \toprule
    & & Math \\

    \midrule

\multirow{6}{*}{Borda} & Diff-token & 63.05 / 67.05 / 68.18 \\
& Diff-token-PreG & \ulgood{64.22} / \ulgood{68.29} / \ulgood{70.00} \\
& Gini & 63.15 / 67.06 / 68.21 \\
& Gini-PreG & \ulgood{64.31} / \ulgood{68.45} / \ulgood{70.09} \\
& Entropy & 63.14 / 67.19 / 68.34 \\
& Entropy-PreG & \ulgood{64.48} / \ulgood{68.65} / \ulgood{70.21} \\
& Conf-Lens & 64.43 / 67.76 / 68.59 \\
& Conf-Lens-PreG & \ulgood{64.85} / \ulgood{69.19} / \ulgood{70.85} \\

\midrule

\multirow{6}{*}{W-10\%} & Diff-token & 57.66 / 57.01 / 62.39 \\
& Diff-token-PreG & \ulgood{58.80} / \ulgood{62.45} / \ulgood{67.52} \\
& Gini & 57.76 / 56.73 / 62.37 \\
& Gini-PreG & \ulgood{59.03} / \ulgood{62.64} / \ulgood{67.49} \\
& Entropy & 58.10 / 57.07 / 62.53 \\
& Entropy-PreG & \ulgood{59.10} / \ulgood{62.90} / \ulgood{67.84} \\
& Conf-Lens & 60.26 / 60.05 / 64.98 \\
& Conf-Lens-PreG & \ulgood{60.63} / \ulgood{64.21} / \ulgood{69.40} \\



    \bottomrule
  \end{tabular}
\end{table}

The results of other voting methods on Qwen are reported in Table~\ref{tab:diff_vote}. The results of other confidence scores methods on Qwen are reported in Table~\ref{tab:other_score}.

\newpage

\clearpage

\section*{NeurIPS Paper Checklist}

\begin{enumerate}

\item {\bf Claims}
    \item[] Question: Do the main claims made in the abstract and introduction accurately reflect the paper's contributions and scope?
    \item[] Answer: \answerYes{} 
    \item[] Justification: The abstract and introduction clearly state that we propose a framework for parallel TTS.
    \item[] Guidelines:
    \begin{itemize}
        \item The answer \answerNA{} means that the abstract and introduction do not include the claims made in the paper.
        \item The abstract and/or introduction should clearly state the claims made, including the contributions made in the paper and important assumptions and limitations. A \answerNo{} or \answerNA{} answer to this question will not be perceived well by the reviewers. 
        \item The claims made should match theoretical and experimental results, and reflect how much the results can be expected to generalize to other settings. 
        \item It is fine to include aspirational goals as motivation as long as it is clear that these goals are not attained by the paper. 
    \end{itemize}

\item {\bf Limitations}
    \item[] Question: Does the paper discuss the limitations of the work performed by the authors?
    \item[] Answer: \answerYes{} 
    \item[] Justification: The limitations are discussed in Section~\ref{sec:conclusion}.
    \item[] Guidelines:
    \begin{itemize}
        \item The answer \answerNA{} means that the paper has no limitation while the answer \answerNo{} means that the paper has limitations, but those are not discussed in the paper. 
        \item The authors are encouraged to create a separate ``Limitations'' section in their paper.
        \item The paper should point out any strong assumptions and how robust the results are to violations of these assumptions (e.g., independence assumptions, noiseless settings, model well-specification, asymptotic approximations only holding locally). The authors should reflect on how these assumptions might be violated in practice and what the implications would be.
        \item The authors should reflect on the scope of the claims made, e.g., if the approach was only tested on a few datasets or with a few runs. In general, empirical results often depend on implicit assumptions, which should be articulated.
        \item The authors should reflect on the factors that influence the performance of the approach. For example, a facial recognition algorithm may perform poorly when image resolution is low or images are taken in low lighting. Or a speech-to-text system might not be used reliably to provide closed captions for online lectures because it fails to handle technical jargon.
        \item The authors should discuss the computational efficiency of the proposed algorithms and how they scale with dataset size.
        \item If applicable, the authors should discuss possible limitations of their approach to address problems of privacy and fairness.
        \item While the authors might fear that complete honesty about limitations might be used by reviewers as grounds for rejection, a worse outcome might be that reviewers discover limitations that aren't acknowledged in the paper. The authors should use their best judgment and recognize that individual actions in favor of transparency play an important role in developing norms that preserve the integrity of the community. Reviewers will be specifically instructed to not penalize honesty concerning limitations.
    \end{itemize}

\item {\bf Theory assumptions and proofs}
    \item[] Question: For each theoretical result, does the paper provide the full set of assumptions and a complete (and correct) proof?
    \item[] Answer: \answerYes{} 
    \item[] Justification: The full proofs are presented in Appendix~\ref{appd:proof}.
    \item[] Guidelines:
    \begin{itemize}
        \item The answer \answerNA{} means that the paper does not include theoretical results. 
        \item All the theorems, formulas, and proofs in the paper should be numbered and cross-referenced.
        \item All assumptions should be clearly stated or referenced in the statement of any theorems.
        \item The proofs can either appear in the main paper or the supplemental material, but if they appear in the supplemental material, the authors are encouraged to provide a short proof sketch to provide intuition. 
        \item Inversely, any informal proof provided in the core of the paper should be complemented by formal proofs provided in appendix or supplemental material.
        \item Theorems and Lemmas that the proof relies upon should be properly referenced. 
    \end{itemize}

    \item {\bf Experimental result reproducibility}
    \item[] Question: Does the paper fully disclose all the information needed to reproduce the main experimental results of the paper to the extent that it affects the main claims and/or conclusions of the paper (regardless of whether the code and data are provided or not)?
    \item[] Answer: \answerYes{} 
    \item[] Justification: We provide full implementation details in Section~\ref{sec:exp}.
    \item[] Guidelines:
    \begin{itemize}
        \item The answer \answerNA{} means that the paper does not include experiments.
        \item If the paper includes experiments, a \answerNo{} answer to this question will not be perceived well by the reviewers: Making the paper reproducible is important, regardless of whether the code and data are provided or not.
        \item If the contribution is a dataset and\slash or model, the authors should describe the steps taken to make their results reproducible or verifiable. 
        \item Depending on the contribution, reproducibility can be accomplished in various ways. For example, if the contribution is a novel architecture, describing the architecture fully might suffice, or if the contribution is a specific model and empirical evaluation, it may be necessary to either make it possible for others to replicate the model with the same dataset, or provide access to the model. In general. releasing code and data is often one good way to accomplish this, but reproducibility can also be provided via detailed instructions for how to replicate the results, access to a hosted model (e.g., in the case of a large language model), releasing of a model checkpoint, or other means that are appropriate to the research performed.
        \item While NeurIPS does not require releasing code, the conference does require all submissions to provide some reasonable avenue for reproducibility, which may depend on the nature of the contribution. For example
        \begin{enumerate}
            \item If the contribution is primarily a new algorithm, the paper should make it clear how to reproduce that algorithm.
            \item If the contribution is primarily a new model architecture, the paper should describe the architecture clearly and fully.
            \item If the contribution is a new model (e.g., a large language model), then there should either be a way to access this model for reproducing the results or a way to reproduce the model (e.g., with an open-source dataset or instructions for how to construct the dataset).
            \item We recognize that reproducibility may be tricky in some cases, in which case authors are welcome to describe the particular way they provide for reproducibility. In the case of closed-source models, it may be that access to the model is limited in some way (e.g., to registered users), but it should be possible for other researchers to have some path to reproducing or verifying the results.
        \end{enumerate}
    \end{itemize}

\item {\bf Open access to data and code}
    \item[] Question: Does the paper provide open access to the data and code, with sufficient instructions to faithfully reproduce the main experimental results, as described in supplemental material?
    \item[] Answer: \answerYes{} 
    \item[] Justification: The code and data are publicly available at
\url{https://github.com/renjie3/PreG},
along with instructions for reproducing the experiments.
    \item[] Guidelines:
    \begin{itemize}
        \item The answer \answerNA{} means that paper does not include experiments requiring code.
        \item Please see the NeurIPS code and data submission guidelines (\url{https://neurips.cc/public/guides/CodeSubmissionPolicy}) for more details.
        \item While we encourage the release of code and data, we understand that this might not be possible, so \answerNo{} is an acceptable answer. Papers cannot be rejected simply for not including code, unless this is central to the contribution (e.g., for a new open-source benchmark).
        \item The instructions should contain the exact command and environment needed to run to reproduce the results. See the NeurIPS code and data submission guidelines (\url{https://neurips.cc/public/guides/CodeSubmissionPolicy}) for more details.
        \item The authors should provide instructions on data access and preparation, including how to access the raw data, preprocessed data, intermediate data, and generated data, etc.
        \item The authors should provide scripts to reproduce all experimental results for the new proposed method and baselines. If only a subset of experiments are reproducible, they should state which ones are omitted from the script and why.
        \item At submission time, to preserve anonymity, the authors should release anonymized versions (if applicable).
        \item Providing as much information as possible in supplemental material (appended to the paper) is recommended, but including URLs to data and code is permitted.
    \end{itemize}

\item {\bf Experimental setting/details}
    \item[] Question: Does the paper specify all the training and test details (e.g., data splits, hyperparameters, how they were chosen, type of optimizer) necessary to understand the results?
    \item[] Answer: \answerYes{} 
    \item[] Justification: We provide full implementation details in Section~\ref{sec:exp}.
    \item[] Guidelines:
    \begin{itemize}
        \item The answer \answerNA{} means that the paper does not include experiments.
        \item The experimental setting should be presented in the core of the paper to a level of detail that is necessary to appreciate the results and make sense of them.
        \item The full details can be provided either with the code, in appendix, or as supplemental material.
    \end{itemize}

\item {\bf Experiment statistical significance}
    \item[] Question: Does the paper report error bars suitably and correctly defined or other appropriate information about the statistical significance of the experiments?
    \item[] Answer: \answerYes{} 
    \item[] Justification: We provide STD in Appendix~\ref{appd:std_results}.
    \item[] Guidelines:
    \begin{itemize}
        \item The answer \answerNA{} means that the paper does not include experiments.
        \item The authors should answer \answerYes{} if the results are accompanied by error bars, confidence intervals, or statistical significance tests, at least for the experiments that support the main claims of the paper.
        \item The factors of variability that the error bars are capturing should be clearly stated (for example, train/test split, initialization, random drawing of some parameter, or overall run with given experimental conditions).
        \item The method for calculating the error bars should be explained (closed form formula, call to a library function, bootstrap, etc.)
        \item The assumptions made should be given (e.g., Normally distributed errors).
        \item It should be clear whether the error bar is the standard deviation or the standard error of the mean.
        \item It is OK to report 1-sigma error bars, but one should state it. The authors should preferably report a 2-sigma error bar than state that they have a 96\% CI, if the hypothesis of Normality of errors is not verified.
        \item For asymmetric distributions, the authors should be careful not to show in tables or figures symmetric error bars that would yield results that are out of range (e.g., negative error rates).
        \item If error bars are reported in tables or plots, the authors should explain in the text how they were calculated and reference the corresponding figures or tables in the text.
    \end{itemize}

\item {\bf Experiments compute resources}
    \item[] Question: For each experiment, does the paper provide sufficient information on the computer resources (type of compute workers, memory, time of execution) needed to reproduce the experiments?
    \item[] Answer: \answerYes{} 
    \item[] Justification: The details of compute resources are provided in Section~\ref{sec:exp}.
    \item[] Guidelines:
    \begin{itemize}
        \item The answer \answerNA{} means that the paper does not include experiments.
        \item The paper should indicate the type of compute workers CPU or GPU, internal cluster, or cloud provider, including relevant memory and storage.
        \item The paper should provide the amount of compute required for each of the individual experimental runs as well as estimate the total compute. 
        \item The paper should disclose whether the full research project required more compute than the experiments reported in the paper (e.g., preliminary or failed experiments that didn't make it into the paper). 
    \end{itemize}
    
\item {\bf Code of ethics}
    \item[] Question: Does the research conducted in the paper conform, in every respect, with the NeurIPS Code of Ethics \url{https://neurips.cc/public/EthicsGuidelines}?
    \item[] Answer: \answerYes{} 
    \item[] Justification: The research conducted in this paper conforms with the NeurIPS Code of Ethics.
    \item[] Guidelines:
    \begin{itemize}
        \item The answer \answerNA{} means that the authors have not reviewed the NeurIPS Code of Ethics.
        \item If the authors answer \answerNo, they should explain the special circumstances that require a deviation from the Code of Ethics.
        \item The authors should make sure to preserve anonymity (e.g., if there is a special consideration due to laws or regulations in their jurisdiction).
    \end{itemize}

\item {\bf Broader impacts}
    \item[] Question: Does the paper discuss both potential positive societal impacts and negative societal impacts of the work performed?
    \item[] Answer: \answerYes{} 
    \item[] Justification: Broader impacts are discussed in Section~\ref{sec:conclusion}.
    \item[] Guidelines:
    \begin{itemize}
        \item The answer \answerNA{} means that there is no societal impact of the work performed.
        \item If the authors answer \answerNA{} or \answerNo, they should explain why their work has no societal impact or why the paper does not address societal impact.
        \item Examples of negative societal impacts include potential malicious or unintended uses (e.g., disinformation, generating fake profiles, surveillance), fairness considerations (e.g., deployment of technologies that could make decisions that unfairly impact specific groups), privacy considerations, and security considerations.
        \item The conference expects that many papers will be foundational research and not tied to particular applications, let alone deployments. However, if there is a direct path to any negative applications, the authors should point it out. For example, it is legitimate to point out that an improvement in the quality of generative models could be used to generate Deepfakes for disinformation. On the other hand, it is not needed to point out that a generic algorithm for optimizing neural networks could enable people to train models that generate Deepfakes faster.
        \item The authors should consider possible harms that could arise when the technology is being used as intended and functioning correctly, harms that could arise when the technology is being used as intended but gives incorrect results, and harms following from (intentional or unintentional) misuse of the technology.
        \item If there are negative societal impacts, the authors could also discuss possible mitigation strategies (e.g., gated release of models, providing defenses in addition to attacks, mechanisms for monitoring misuse, mechanisms to monitor how a system learns from feedback over time, improving the efficiency and accessibility of ML).
    \end{itemize}
    
\item {\bf Safeguards}
    \item[] Question: Does the paper describe safeguards that have been put in place for responsible release of data or models that have a high risk for misuse (e.g., pre-trained language models, image generators, or scraped datasets)?
    \item[] Answer: \answerNA{} 
    \item[] Justification: This paper does not release any models or datasets that pose a high risk for misuse.
    \item[] Guidelines:
    \begin{itemize}
        \item The answer \answerNA{} means that the paper poses no such risks.
        \item Released models that have a high risk for misuse or dual-use should be released with necessary safeguards to allow for controlled use of the model, for example by requiring that users adhere to usage guidelines or restrictions to access the model or implementing safety filters. 
        \item Datasets that have been scraped from the Internet could pose safety risks. The authors should describe how they avoided releasing unsafe images.
        \item We recognize that providing effective safeguards is challenging, and many papers do not require this, but we encourage authors to take this into account and make a best faith effort.
    \end{itemize}

\item {\bf Licenses for existing assets}
    \item[] Question: Are the creators or original owners of assets (e.g., code, data, models), used in the paper, properly credited and are the license and terms of use explicitly mentioned and properly respected?
    \item[] Answer: \answerYes{} 
    \item[] Justification: All datasets and models used in this paper are properly cited. The licenses are presented in Appendix~\ref{appd:license}.
    \item[] Guidelines:
    \begin{itemize}
        \item The answer \answerNA{} means that the paper does not use existing assets.
        \item The authors should cite the original paper that produced the code package or dataset.
        \item The authors should state which version of the asset is used and, if possible, include a URL.
        \item The name of the license (e.g., CC-BY 4.0) should be included for each asset.
        \item For scraped data from a particular source (e.g., website), the copyright and terms of service of that source should be provided.
        \item If assets are released, the license, copyright information, and terms of use in the package should be provided. For popular datasets, \url{paperswithcode.com/datasets} has curated licenses for some datasets. Their licensing guide can help determine the license of a dataset.
        \item For existing datasets that are re-packaged, both the original license and the license of the derived asset (if it has changed) should be provided.
        \item If this information is not available online, the authors are encouraged to reach out to the asset's creators.
    \end{itemize}

\item {\bf New assets}
    \item[] Question: Are new assets introduced in the paper well documented and is the documentation provided alongside the assets?
    \item[] Answer: \answerYes{} 
    \item[] Justification: The generated data, code, and documentation
are publicly available at
\url{https://github.com/renjie3/PreG}.
    \item[] Guidelines:
    \begin{itemize}
        \item The answer \answerNA{} means that the paper does not release new assets.
        \item Researchers should communicate the details of the dataset\slash code\slash model as part of their submissions via structured templates. This includes details about training, license, limitations, etc. 
        \item The paper should discuss whether and how consent was obtained from people whose asset is used.
        \item At submission time, remember to anonymize your assets (if applicable). You can either create an anonymized URL or include an anonymized zip file.
    \end{itemize}

\item {\bf Crowdsourcing and research with human subjects}
    \item[] Question: For crowdsourcing experiments and research with human subjects, does the paper include the full text of instructions given to participants and screenshots, if applicable, as well as details about compensation (if any)? 
    \item[] Answer: \answerNA{} 
    \item[] Justification: This paper does not involve crowdsourcing experiments or research with human subjects.
    \item[] Guidelines:
    \begin{itemize}
        \item The answer \answerNA{} means that the paper does not involve crowdsourcing nor research with human subjects.
        \item Including this information in the supplemental material is fine, but if the main contribution of the paper involves human subjects, then as much detail as possible should be included in the main paper. 
        \item According to the NeurIPS Code of Ethics, workers involved in data collection, curation, or other labor should be paid at least the minimum wage in the country of the data collector. 
    \end{itemize}

\item {\bf Institutional review board (IRB) approvals or equivalent for research with human subjects}
    \item[] Question: Does the paper describe potential risks incurred by study participants, whether such risks were disclosed to the subjects, and whether Institutional Review Board (IRB) approvals (or an equivalent approval/review based on the requirements of your country or institution) were obtained?
    \item[] Answer: \answerNA{} 
    \item[] Justification: This paper does not involve research with human subjects.
    \item[] Guidelines:
    \begin{itemize}
        \item The answer \answerNA{} means that the paper does not involve crowdsourcing nor research with human subjects.
        \item Depending on the country in which research is conducted, IRB approval (or equivalent) may be required for any human subjects research. If you obtained IRB approval, you should clearly state this in the paper. 
        \item We recognize that the procedures for this may vary significantly between institutions and locations, and we expect authors to adhere to the NeurIPS Code of Ethics and the guidelines for their institution. 
        \item For initial submissions, do not include any information that would break anonymity (if applicable), such as the institution conducting the review.
    \end{itemize}

\item {\bf Declaration of LLM usage}
    \item[] Question: Does the paper describe the usage of LLMs if it is an important, original, or non-standard component of the core methods in this research? Note that if the LLM is used only for writing, editing, or formatting purposes and does \emph{not} impact the core methodology, scientific rigor, or originality of the research, declaration is not required.
    \item[] Answer: \answerNA{} 
    \item[] Justification: LLMs were only used for writing assistance and do not form part of the core methodology.
    \item[] Guidelines:
    \begin{itemize}
        \item The answer \answerNA{} means that the core method development in this research does not involve LLMs as any important, original, or non-standard components.
        \item Please refer to our LLM policy in the NeurIPS handbook for what should or should not be described.
    \end{itemize}

\end{enumerate}

\end{document}